\documentclass[10pt,journal,cspaper,compsoc]{IEEEtran}

\ifCLASSOPTIONcompsoc
  \usepackage[nocompress]{cite}
\else
  \usepackage{cite}
\fi

\ifCLASSINFOpdf
   \usepackage[pdftex]{graphicx}
   \graphicspath{{../pdf/}{../jpeg/}}
   \DeclareGraphicsExtensions{.pdf,.jpeg,.png,.jpg}
\else
   \usepackage[dvips]{graphicx}
   \graphicspath{{../eps/}}
   \DeclareGraphicsExtensions{.eps}
\fi

\usepackage{amsmath}
\usepackage{amsfonts}

\ifCLASSOPTIONcompsoc
  \usepackage[caption=false,font=footnotesize,labelfont=sf,textfont=sf]{subfig}
\else
  \usepackage[caption=false,font=footnotesize]{subfig}
\fi

\usepackage{stfloats}

\usepackage{amsmath}
\usepackage{amsfonts}
\usepackage{enumerate}

\def\ie{{\em i.e.}}
\def\eg{{\em e.g.}}
\def\etal{{\em et al.}}

\newcommand{\figref}[1]{Fig.~\ref{#1}}
\newcommand{\tabref}[1]{Tab.~\ref{#1}}

\newcommand{\secref}[1]{Section \ref{#1}}

\newcommand{\mc}[1]{\mathcal{#1}}

\newcommand{\br}[1]{\bm{\mathrm{#1}}}

\newcommand{\bs}[1]{\boldsymbol{\texttt{#1}}}

\usepackage{ragged2e}
\usepackage{booktabs}
\usepackage{color}
\usepackage{multirow}
\usepackage{bm}
\usepackage{bbm}

\usepackage{xcolor}
\usepackage{balance}

\usepackage{amssymb}

\usepackage[noend]{algpseudocode}
\usepackage{algorithmicx,algorithm}

\graphicspath{{./figures/}}

\begin{document}

\title{
Language-Inspired Relation Transfer for Few-shot Class-Incremental Learning
}

\author{Yifan~Zhao,~\IEEEmembership{Member,~IEEE},~Jia~Li,~\IEEEmembership{Senior Member,~IEEE}, ~Zeyin~Song,
and~Yonghong~Tian,~\IEEEmembership{Fellow,~IEEE}% <-this % stops a space
\IEEEcompsocitemizethanks{
\IEEEcompsocthanksitem Y. Zhao and J. Li are with the State Key Laboratory of Virtual Reality Technology and Systems, School of Computer Science and Engineering, Beihang University, Beijing, 100191, China.
\IEEEcompsocthanksitem Z. Song and Y. Tian are with the School of Electronic and Computer Engineering, Peking University, Shenzhen, 518055, China.
\IEEEcompsocthanksitem  Y. Tian is with the School of Computer Science, Peking University, Beijing, 100871, China
\IEEEcompsocthanksitem Y. Tian is also with the Pengcheng Laboratory, Shenzhen, 518055, China.

\IEEEcompsocthanksitem J. Li and Y. Tian are the corresponding authors (E-mail: jiali@buaa.edu.cn, yhtian@pku.edu.cn).
}}

\markboth{Submission to IEEE TRANSACTIONS ON PATTERN ANALYSIS AND MACHINE INTELLIGENCE}%
{Shell \MakeLowercase{\textit{et al.}}:}

\IEEEtitleabstractindextext{%
\begin{abstract}
\justifying Depicting novel classes with language descriptions by observing few-shot samples is inherent in human-learning systems. This lifelong learning capability helps to distinguish new knowledge from old ones through the increase of open-world learning, namely Few-Shot Class-Incremental Learning (FSCIL). Existing works to solve this problem mainly rely on the careful tuning of visual encoders, which shows an evident trade-off between the base knowledge and incremental ones. Motivated by human learning systems, we propose a new Language-inspired Relation Transfer (LRT) paradigm to understand objects by joint visual clues and text depictions, composed of two major steps. We first transfer the pretrained text knowledge to the visual domains by proposing a graph relation transformation module and then fuse the visual and language embedding by a text-vision prototypical fusion module. Second, to mitigate the domain gap caused by visual finetuning, we propose context prompt learning for fast domain alignment and imagined contrastive learning to alleviate the insufficient text data during alignment. With collaborative learning of domain alignments and text-image transfer, our proposed LRT outperforms the state-of-the-art models by over $13\%$ and $7\%$ on the final session of \textit{mini}ImageNet and CIFAR-100 FSCIL benchmarks. 
\end{abstract}

}

\maketitle

\IEEEdisplaynontitleabstractindextext

\IEEEpeerreviewmaketitle

\IEEEraisesectionheading{\section{Introduction}\label{sec:introduction}}

\IEEEPARstart{H}{uman} brains show their distinctive advantages in recognizing new concepts with only a few limited samples, while not forgetting the old knowledge rapidly. Benefited from the strong perceptual capability of deep neural networks~\cite{he2016deep,radford2021learning}, recent advances propose to imitate human learning systems mainly from two aspects,~\ie, recognizing new concepts with extremely few samples and learning without forgetting. For the first few-shot learning (FSL) challenge,
existing works focus on network learning with fast optimization strategies~\cite{li2017meta,rezende2016one,santoro2016meta} or measuring with appropriate metrics~\cite{snell2017prototypical,sung2018learning,vinyals2016matching}. And to solve the second challenge as well as alleviate forgetting, class-incremental learning (CIL) methods have made significant progress with mechanisms including rehearsal~\cite{rebuffi2017icarl,chaudhry2018efficient,rolnick2019experience}, novel model consolidation~\cite{zhang2020class,mallya2018packnet} and feature space regularization strategies~\cite{li2017learning}. Nevertheless, when considering these two natural abilities together, unlike human-learning systems, existing methods encounter significant obstacles~\cite{tao2020few} for generalizing on new concepts or catastrophic forgetting on base knowledge due to the limited new samples for training.

\begin{figure}[!t]
	\centering
	\includegraphics[width=\columnwidth]{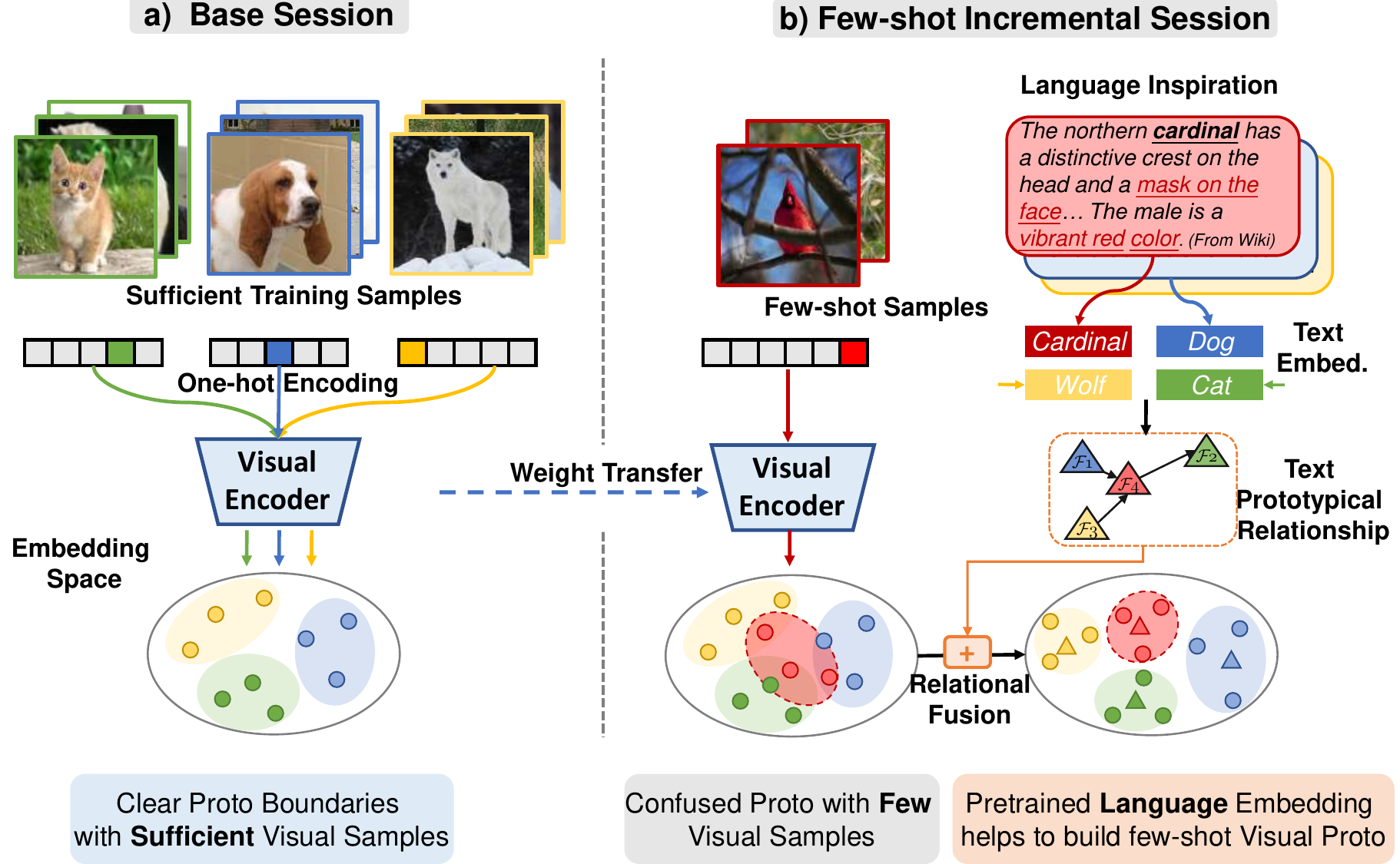}
	\caption{The motivation of the proposed approach. Visual encoders provide clear boundaries in a) when learning with base sufficient data, while resulting in confused prototypes with only a few samples of novel classes in b). Our proposed LRT aims to transfer the pretrained language relationships to help construct a joint feature representation of both base and novel classes.
	}
	\label{fig:motivation}
\end{figure}

One intuitive idea to solve this problem,~\ie, few-shot class-incremental learning (FSCIL), is to adopt knowledge distillation~\cite{cheraghian2021semantic,dong2021few} from base classes when gradually learning new concepts. As only a few samples are accessible during incremental phases, naive distillations with these seen samples also lead to severe overfitting. To alleviate this, prevailing works dedicate to decoupling base and incremental learning stages~\cite{zhang2021few} and then fix or slightly tune the backbone representations~\cite{mazumder2021few,chen2020incremental,hersche2022constrained}. Besides these works, other research efforts tend to find generalized feature representations by using sufficient training data from base sessions. Representative works propose to find the flattened region in optimization~\cite{shi2021overcoming} or construct virtual classes~\cite{zhou2022forward} when sufficient training samples are available.
Although these works tried to achieve a balanced performance trade-off between the base classes and incremental classes, the dilemma still exists: how to represent the novel incremental classes well without losing distinguishability on the base classes?

When sufficient training samples are available, supervised learning systems present superior performances with state-of-the-art visual encoders. As in~\figref{fig:motivation} a), visual prototypes of base seen classes filled the embedding space and show clear classification boundaries. However, in~\figref{fig:motivation} b), during incremental sessions, features of new classes are still represented with the identical encoder that is trained with base classes, which thus leads to prototype confusion or topological damages. In this paper, we argue to solve this dilemma by a new Language-inspired Relation Transfer (LRT) paradigm, which is motivated by the recent advances of Contrastive Language-Image Pretraining (CLIP)~\cite{radford2021learning}. Different from prevailing methods with static visual embedding, when learning a novel category of \textit{Cardinal Bird} in~\figref{fig:motivation} b), we introduce the language prompt (several words or description sentences) as auxiliary information if there are no sufficient visual samples. Besides, this contrastive learning paradigm constructs a unified feature alignment space of text prompts and image-level features. Hence to transfer the text knowledge, our method first builds a graph relation transformation module to transfer the well-embedded language relationships to the inferior visual space, thus the tangled visual features can be reprojected in the correct space driven by the strong language guidance.

Recent trends to predict object classes in CLIP-based models~\cite{radford2021learning,li2022grounded,zhou2022learning} is to calculate the similarity between text and image embedding. Although this zero-shot trend provides preferable generalization capabilities on new classes, it remains a gap~\cite{wortsman2022robust} compared to the performances using fully supervised visual models. Combining the merits of supervised vision models and language-vision contrastive relations, we consolidate text embedding of one category for prototypical representation and then propose a text-vision prototypical fusion module to incorporate representations from both visual and text domains. In this way, the well-trained language embedding provides strong backing for the standard visual representations, especially in data-deficient few-shot scenarios.
However, only finetuning visual data would lead to a misalignment of visual and language domains. Thus we first introduce a context text prompt learning module to depict few-shot visual samples with learnable text prompts instead of the hand-crafted ones in vanilla CLIP~\cite{radford2021learning}, which fast mitigates the domain gap with only few incremental samples.

Beyond these improvements in the knowledge transfer module, we also notice that the multi-modality contrastive training would be easy to overfit on specific data domains. It is because although the image visual data are various and sufficient, its corresponding language descriptions (\ie,   label texts in our approach) are \textit{monotonous}. Thus to solve this brand new problem, for the multimodal alignment, we randomly mix the input images and also mix their text labels including the learnable prompt tokens as a virtual class. Then the imagined contrastive learning is conducted among these \textit{imagined} prototypes and theoretically $N$ times ($N$ is the number of classes) larger than the vanilla text input space.
With the collaboration of text-to-image relation transfer and multi-modal alignment, our proposed LRT is able to achieve a comprehensive understanding of one novel concept without forgetting the old ones. Moreover, LRT does not rely on any auxiliary networks (including the text encoder) during the inference time, making the final model lightweight and implementation-friendly. Experimental evidence demonstrates that LRT outperforms the state-of-the models by $13.3\%$ on \textit{mini}ImageNet~\cite{deng2009imagenet} and $7.3\%$ on CIFAR-100~\cite{krizhevsky2009learning} benchmarks in the final session.

In summary, our contribution is threefold:
1) We make an attempt to solve the few-shot class-incremental learning with pretrained language understanding and propose a new Language-inspired Relation Transfer (LRT) paradigm.
2) We propose a graph relation transformation module to gradually transfer the text knowledge into few-shot visual prototypes, and introduce a text-vision prototypical fusion strategy for feature representation, which combines the merits of the visual embedding and pretrained language guidance.
3) We propose a context text prompt learning strategy to align the text and image domains with few shots and an imagined contrastive learning strategy to alleviate the \textit{insufficient text} label spaces for generalization representation.

The remainder of this paper is organized as follows:~\secref{sec:relatedwork} reviews related works and discusses the relations among previous literature. ~\secref{sec:method} describes the proposed language-inspired relation transfer approach. Qualitative and quantitative experiments with detailed analyses are exhibited in~\secref{sec:exp} and~\secref{sec:conclusion} finally concludes this paper.

\section{Related Work}\label{sec:relatedwork}
\textbf{Few-shot Learning.} Inspired by human recognition systems, few-shot learning aims to distinguish conceptually new object categories by inferring from base knowledge. Recent ideas to solve this problem could be roughly divided into two trends: model optimization~\cite{li2017meta, rezende2016one,santoro2016meta,ravi2017optimization,finn2017model,nichol2018first} and metric learning manners~\cite{triantafillou2017few,ren2018meta,oreshkin2018tadam,snell2017prototypical,sung2018learning,vinyals2016matching}. Optimization-based methods focuses on the generalization ability by using meta-learning frameworks. For example, model-agonistic meta-learning~\cite{finn2017model,nichol2018first} aims to learn the fast adaptation ability by learning from the direction of sampled task gradients. While metric-learning-based methods focus on the distance measurement of novel query samples and base knowledge representations. Representative works focuses on the prototype learning~\cite{snell2017prototypical}, local representations~\cite{wertheimer2021few} and feature space reprojections~\cite{zhang2020deepemd}.

\textbf{Class-incremental Learning.} Class-Incremental Learning (CIL) focuses on one specific direction of the field of continual learning~\cite{parisi2019continual}, which aims to learn from new classes without forgetting the base knowledge. Prevailing research dedicated to this task focuses on replaying the old memories ~\cite{rebuffi2017icarl,chaudhry2018efficient,rolnick2019experience,Belouadah_2019_ICCV,Hu_2021_CVPR} and regularizing the feature space~\cite{li2017learning,liu2018rotate,tao2020topology}. Representative methods in the first family including iCaRL~\cite{rebuffi2017icarl}, CLEAR~\cite{rolnick2019experience} and A-GEM~\cite{chaudhry2018efficient} selectively retain the knowledge from old samples and replay these samples or features when learning the new classes. For example,  iCaRL~\cite{rebuffi2017icarl} aims to distill the base knowledge when learning samples from new categories, which greatly alleviates catastrophic forgetting. While the second family of methods~\cite{li2017learning,liu2018rotate} tends to build regularized feature space and
Besides these with fixed model structures, the other line of works proposes to solve this problem by model ensemble~\cite{zhang2020class} and iterative pruning~\cite{mallya2018packnet}.
This research direction also shares common concerns with few-shot learning to represent new classes. However, when tackling incremental categories with very few samples, rehearsal or distillation-based methods usually face severely catastrophic overfitting and fail to represent the novel categories.

\textbf{Few-shot Class-incremental Learning.} As a newly proposed realistic setting, Few-Shot Class-Incremental Learning (FSCIL) proposed by~\cite{tao2020few} has attracted considerable attention. Inspired by incremental learning methods, several research~\cite{cheraghian2021semantic,dong2021few} propose to alleviate the forgetting of base classes by knowledge distillation during few-shot learning. Zhao~\etal~\cite{zhao2021mgsvf} propose a slow-fast updating framework to achieve a balanced trade-off between the novel updating and old knowledge degradation.
As the base samples during incremental learning are infeasible, prevailing methods~\cite{shi2021overcoming,zhou2022forward,zoumargin,peng2022few,zhou2022few} tend to find the generalized representation during base sessions. For example, Zhou~\etal~\cite{zhou2022few} propose to synthesize fake FSCIL tasks from the base dataset with meta-learning strategies.   
Besides, other works propose to resist the overfitting caused by insufficient training samples by using graph models~\cite{zhang2021few} or selected parameter adjustment~\cite{mazumder2021few,chen2020incremental,hersche2022constrained}. Hersche~\etal~\cite{hersche2022constrained} design a semi-frozen meta-learning framework with rewritable dynamically growing memory. However, although the discovery abilities of novel categories are improved, the frozen visual backbones still restrict their representation abilities to extract sufficient visual cues. 

\textbf{Contrastive Vision-Language Model.} Cross-modality pretraining with self-supervised contrastive learning has been widely adopted in various applications. Representative vision-language models including CLIP~\cite{radford2021learning}, ALIGN~\cite{jia2021scaling} and CyCLIP~\cite{goel2022cyclip} have shown great success in zero-shot image recognition tasks. Inspired by these works, contrastive pretraining using multi-view~\cite{li2021supervision} or part-level supervision~\cite{li2022grounded} has enlightened many down-stream vision tasks,~\eg, zero-shot object detection and visual question answering.
Moreover, several very recent works focus on prompt engineering to make a fast adaptation on target domains, including vision prompt~\cite{wang2022learning} and language prompt~\cite{zhou2022learning}. Although these aforementioned methods show effectiveness in zero-shot learning, as validated in~\cite{zhou2022learning,wortsman2022robust}, there is still a huge gap between supervised learning and CLIP-based models. In addition, when jointly optimizing these models, the base classes and novel categories show less distribution gap which cannot be jointly optimized in the few-shot class-incremental setting.

\textbf{Discussions and Relations.} Methods of few-shot learning and class-incremental learning only focus on the single side of the FSCIL problem. Prevailing few-shot class-incremental learning methods achieve preferable performance by alleviating the catastrophic overfitting of base sessions, while the novel discovery capability is still restricted by the inferior representation features trained by limited samples. To overcome this bottleneck, in this paper, we argue that one promising solution to understanding few-shot objects with incremental ability is from the generalized visual-language knowledge: 1) fast adapting the generalized representation to downstream task-specific features, 2) excavating generalized language knowledge to guide the learning of few-shot visual samples, and 3) maintaining the text-image cross-modal alignment with only few samples.

\begin{figure}
\begin{center}
\includegraphics[width=1\columnwidth]{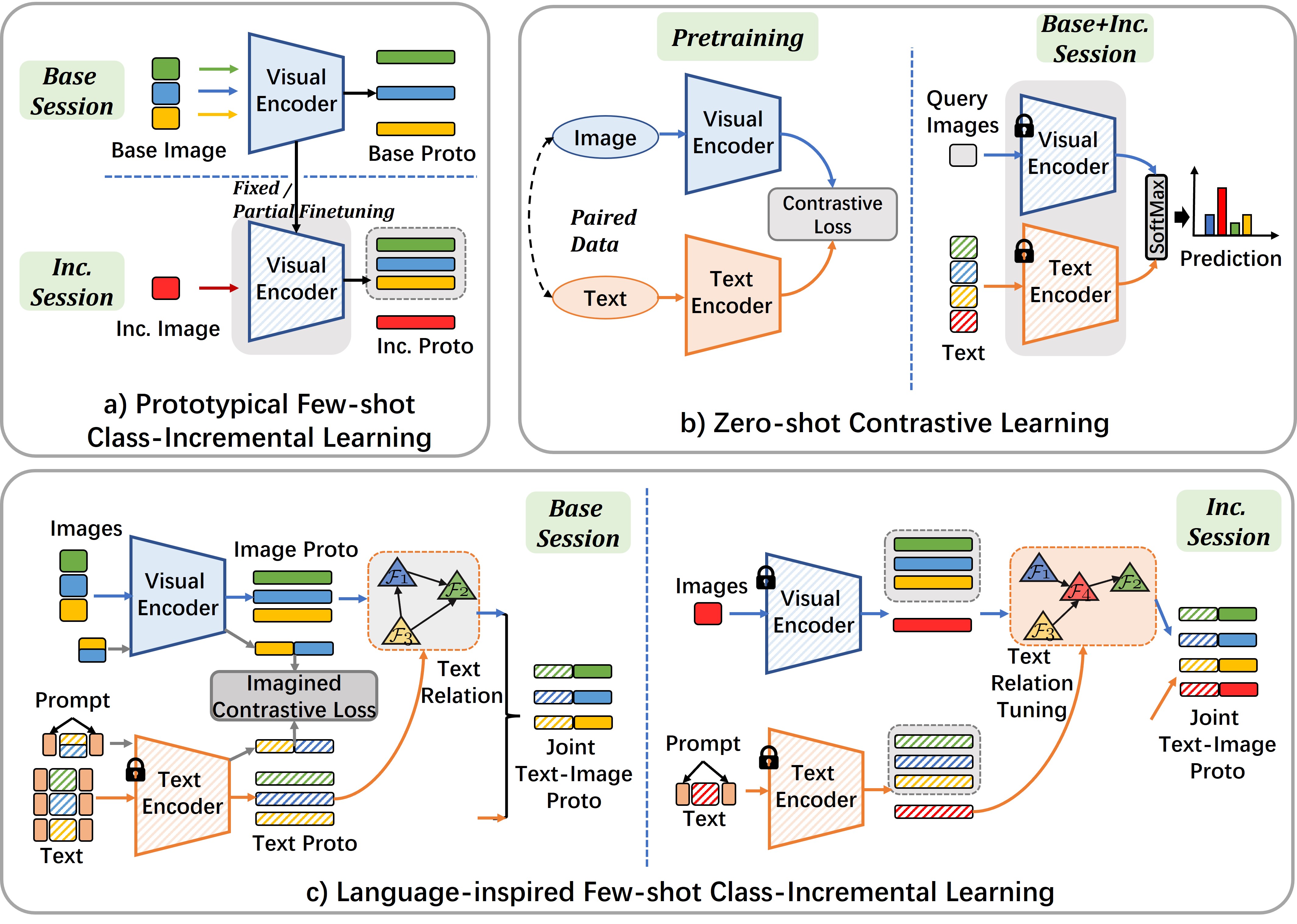}
 \caption{Illustrations of different learning paradigms. a) Prototypical FSCIL~\cite{zhang2021few,zhou2022forward}: using visual prototypes for incremental classes. b) Zero-shot CLIP~\cite{radford2021learning}: direct predicting probabilities after image-text contrastive learning.
 c) Ours: transferring the pretrained text embedding to visual domains meanwhile keeping domain alignment with context prompt and imagined contrastive loss.
 }\label{fig:comp}
 \end{center}

\end{figure}

\begin{figure*}
\begin{center}
\includegraphics[width=0.95\textwidth]{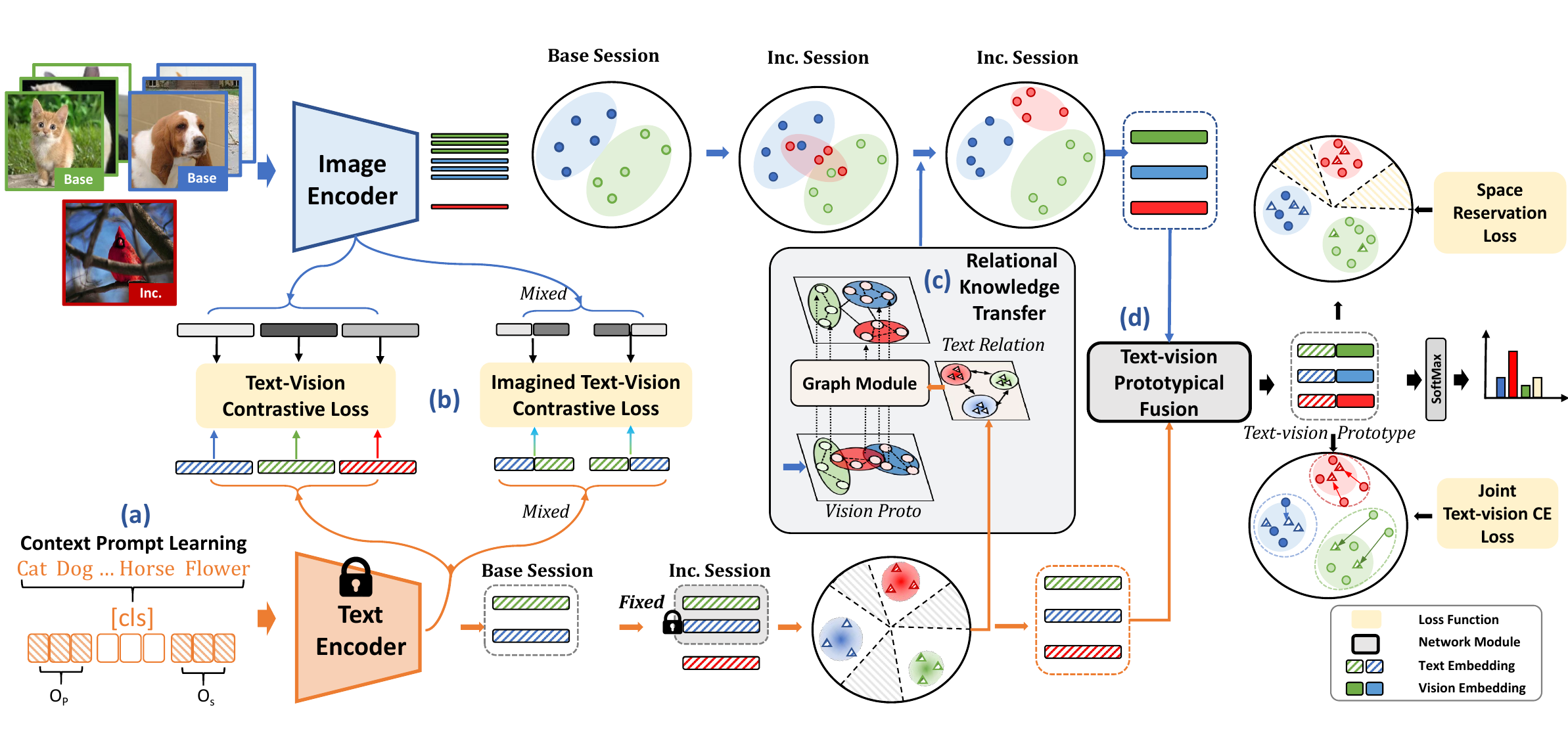}
 \caption{The proposed Language-inspired Relation Transfer (LRT) approach consists of two essential modules. 1) Relational knowledge transfer module first transfers the text-wise relationship to the visual prototypes and a text-vision prototypical fusion module for knowledge fusion. 2) Image-Text alignment module introduces context prompt learning for fast adaptation and proposes the imagined contrastive learning for multi-modal alignment in few-shot class incremental learning. }
 \label{fig:pipeline}
 \end{center}
\end{figure*}

\section{Approach}\label{sec:method}

\subsection{Problem Formulations and Baselines}\label{sec:pre}
\textbf{Few-shot Class-incremental Learning with Text.} FSCIL focuses on the intersection of class-incremental learning and few-shot learning problems, which aims to jointly recognize the incremental classes and base classes with a sequential of given sessions. An FSCIL model sequentially receives $S$ training session $\mc{D}^1 \dots \mc{D}^S$ with sets of triplets.~\ie, $\mc{D}^{s}=\{(\br{x}_{i}^{s},\br{y}_{i}^{s},\br{t}_{i}^{s})\}_{i=1}^{|\mc{D}^{s}|}$, where $\br{x}_{i}^{s} \in \mc{X}^{s}, (\br{y}_{i}^{s},\br{t}_{i}^{s}) \in \mc{C}^{s} \times \mc{E}^{s}$ denotes the training images, one-hot labels, and text labels with class names respectively. $\mc{X}, \mc{C}$ and $\mc{E}$ are space notations for the visual, label, and text domains. During the training of FSCIL, the base session $\mc{D}^1$ contains sufficient training samples of base classes, and the subsequent $2\sim S$ sessions are defined as typical N-way M-shot few-shot learning problems. During the incremental session, only samples in the current session are visible and label spaces do not contain any overlap. $\mc{C}^{i} \cap \mc{C}^{j} = \varnothing, \forall i,j\in \{1 \ldots S\}, i\ne j$, and similarly we have $\mc{D}^{i} \cap \mc{D}^{j} = \varnothing, \mc{E}^{i} \cap \mc{E}^{j} = \varnothing$. With learnable parameters $\Theta$, the overall learning objective is to minimize the measurement $\xi$ across all sessions:
\begin{equation}\label{eq:defination}
\arg\min_{\Theta}\Sigma_{s=1}^{S}\Sigma_{(\br{x},\br{y},\br{t})\sim \mc{D}^{s}}\xi(f_{\Theta}(\br{x};\br{t}),\br{y}),
\end{equation}
where $\xi$ are usually set as cosine or Euclidean distances with the one-hot class label \br{y} and the label text $\br{t}$ for each class are introduced as auxiliary input. 

\textbf{Visual Learning Baseline.} One intuitive but effective visual learning scheme recently~\cite{zhang2021few,zhou2022forward} is to use prototypical networks~\cite{snell2017prototypical} both for base and incremental sessions. We denote $\mc{V}_{B},\mc{V}_{I}$ for visual encoders of base and incremental sessions respectively. The prototypical networks rely on the slowly updated or fixed visual encoder $\mc{V}_{B}$ that is pretrained on base sessions. During the incremental session, the weight of base visual encoder is transferred to the incremental visual encoder $\mc{V}_{I}(\cdot)\leftarrow \mc{V}_{B}(\cdot)$ (fixed or slightly tuned). The fully connected layers for classifiers are replaced with feature prototypes, as in~\figref{fig:comp} a). Thus the visual prototypes $\br{V}^{i}$ across all classes have the form:
\begin{equation}\label{eq:visualproto}
\br{V}^{i}=\frac{1}{WH}\sum_{(\br{x},\br{y}) \sim \mc{D},\br{y}=\mc{C}_{i}} \sum_{j=1}^{WH}    \mc{V}_{\{B,I\}}(\br{x}_{j}; \mc{C}_{i}),
\end{equation}
where $W,H$ denote the width and height of feature maps respectively. With the averaged prototypes of all classes $\{\br{V}^{i}\}_{i=1}^{|\mc{C}|}\in \mathbb{R}^{1\times 1 \times D_{V}}$ and the standard measurements $\xi(\br{x},\br{V})$, visual models show strong capabilities in alleviating \textit{catastrophic forgetting}. Nevertheless, they are easy to overfit on the limited few-shot incremental data and cannot form the \textit{generalized embedding}.

\subsection{Connecting Images with Texts in FSCIL}\label{sec:t2i}
\begin{figure}[!t]
	\centering
	\includegraphics[width=\columnwidth]{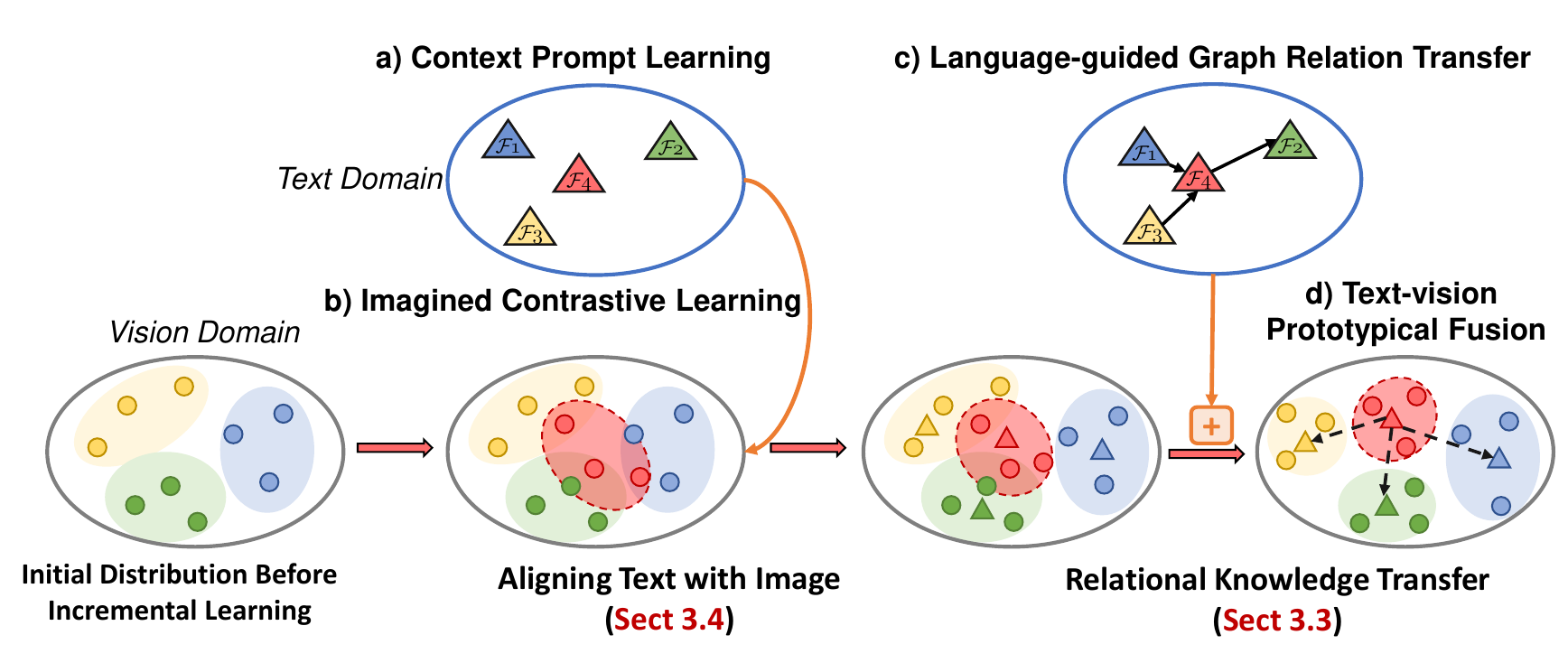}
	\caption{The motivation and modules of the proposed LRT. Our proposed LRT is composed of an aligning stage to conduct a multimodal alignment with few-shot downstream data and a transferring stage to transfer the text knowledge to the vision domain.}
	\label{fig:overview}
\end{figure}

\textbf{Language-guided FSCIL Paradigm.} 
Our main motif is to utilize the pretrained knowledge in the text domain to facilitate the learning of few-shot class incremental sessions. To achieve this, we face two major dilemmas beyond the prevailing incremental learning challenges, 1) visual representation scarcity of novel concepts and 2) continual misalignment of multi-modalities caused by imbalanced and downstream learning tasks. 
Toward these dilemmas, our major pipeline can be simplified as two major steps as in~\figref{fig:overview}, ~\ie, \textbf{transferring} and \textbf{aligning}. For the first dilemma, we advocate transferring the pretrained generalized language concept knowledge to the visual modality by relational knowledge transfer module in \secref{sec:relation}. Note that this module is constructed for both the base and incremental learning sessions. For the second misalignment dilemma, in \secref{sec:align}, we propose the imagined aligning strategy for the base pretraining session and context prompt adaptation only for the incremental session, which jointly alleviate severely misaligned text and visual modality during learning.

\textbf{Zero-shot Measurements with Texts.} Contrastive pretraining vision-language models including CLIP~\cite{radford2021learning} and ALIGN~\cite{jia2021scaling}, have offered us a conceptually new solution to solve the few-shot representation predicament. As in~\figref{fig:comp} b), taking the advantages of rich language data, this contrastive learning trend shows significant \textit{generalization} ability on extremely few-shot image samples. Given a cluster of $N$ text labels to predict,~\ie,$\{(\br{t}, \br{y})| \br{y}=\mc{C}_i\}_{i=1}^{N}$ , the text encoder $\mc{T}(\cdot)$ aligns the text inputs and the image features of query $\br{x}$ in the same space. Hence the zero-shot prediction of each class $\mc{C}_i$ is presented as:
\begin{equation}\label{eq:clip}
\br{p}_{i}=\frac{e^{\xi(\mc{V}(\br{x}),\mc{T}(\br{t}_{i})^{\top})}}{\sum_{j=i}^{N} e^{\xi(\mc{V}(\br{x}),\mc{T}(\br{t}_{j})^{\top})}},
\end{equation}
where $\xi(\br{x},\br{t}) = {\br{x}\cdot \br{t}} / (||\br{x}||^{2}_{2} ||\br{t}||^{2}_{2})$ denotes the normalized cosine similarity with omitted scale factors for simplicity. However, this equation only measures the similarity of input images with text prototypes, while omitting the similarity of input to image prototypes.

\subsection{Relational Knowledge Transfer} \label{sec:relation}
Although the CLIP-based models show strong generalization capability on unseen categories, FSCIL faces a conceptually different problem,~\ie, sufficient visual samples of base classes are available and few-shot incremental samples in the same vein. Omitting these visual clues as well as the class-based prototypes would lead to overfitting due to the target dataset being small compared to the large pretraining domain. Besides, the language-vision pretraining models show a clear performance drop compared with the supervised learning, as demonstrated in~\cite{wortsman2022robust,radford2021learning}. Toward this end, we propose constructing two major modules for knowledge transfer, ~\ie, the language-guided graph relation transfer (\figref{fig:overview} c)) and text-vision prototypical fusion (\figref{fig:overview} d)). These two modules are consistently constructed for both the base pretraining session and the subsequent incremental learning session.

\textbf{Language-guided Graph Relation Transfer.} Inspired by the prototypical learning, here we adopt the text encoding features of the same class $\br{T}_{i}= 1/K \sum_{k}\mc{T}(\br{t}_k),\forall \br{y}_k =\mc{C}_i$ as the \textbf{text prototypes} to represent the features of class $\mc{C}_i$, where K denotes the number of text prompts. This embedding can be formed by using class names or even incorporating the object context features,~\eg, \textit{``cardinals are usually in red color"} in~\figref{fig:motivation}, which can provide rich prior knowledge for object recognition, especially in few-shot scenarios. More importantly, the generated text prototypes are naturally distributed in the same space as the visual features and benefited from the contrastive multi-modal pretraining. Considering~\figref{fig:pipeline}, the most crucial challenge in incremental sessions is that the new visual prototypes are entangled with the base ones. We therefore decide to disentangle these confused samples by introducing the relationship from pretrained language domain. The pair-wise relationship of text prototypes $\mc{T}(\br{t}_{i}) \in \mathbb{R}^{1\times 1\times D_{T}}$ is:
\begin{equation} \label{eq:adj}
\br{A}_{i,j} = \frac{\mc{T}(\br{t}_{i})^{\top} \cdot \mc{T}(\br{t}_{j})} {\left\|\mc{T}(\br{t}_{i}) \right\| \left\| \mc{T}(\br{t}_{j})\right\|}.
\end{equation}

We then construct a relationship transformation graph with the visual prototypes as graph nodes,~\ie, $\mc{G}=\{\br{V},\br{A}\}, \br{V}=\{\mc{V}(\br{x})\}_{i=1}^{|\mc{C}|}$ and the $\mc{C}$ can denote base classes $\mc{C}^{base}$ or  $[\mc{C}^{base},\mc{C}^{inc}]$ during the incremental session. With the relation adjacent matrix, the reprojected visual prototypes using graph convolutional networks~\cite{welling2016semi} in~\figref{fig:pipeline} is formally presented as:
\begin{equation} \label{eq:gcn}
\br{U} = \bs{ReLU}(\br{\widetilde{D}}^{-\frac{1}{2}} \br{\widetilde{A}} \br{\widetilde{D}}^{-\frac{1}{2}} \br{V} \br{W}^v )\in \mathbb{R}^{|\mc{C}|\times{D_V}},
\end{equation}
where $\br{W}^v \in \mathbb{R}^{D_V \times D_V}$ is the learnable graph weights. Here we set the output dimensions of the text and visual features are aligned $D_V=C_T$ for subsequent fusion operations. $\br{\widetilde{D}}=\sum_j\br{\widetilde{A}_{i,j}}$ is the normalized diagonal matrix. $\br{\widetilde{A}}= \br{A}+\br{I} \in \mathbb{R}^{|\mc{C}| \times |\mc{C}|}$ denotes the text relationship with self-loop and $\br{I} $ denotes the identity matrix.

\textbf{Text-vision Prototypical Fusion.} With the graph relation transferring from text features, the updated visual prototypes $\br{U}$ are reprojected in a topologically distinguishable space for recognition. In this paper, we argue that text prototypes in~\figref{fig:comp} b) and visual prototypes in a) are both beneficial for FSCIL tasks,~\ie, the text prototypes provide well-generalized representations when there are insufficient training samples, meanwhile, the visual prototypes provide clear visual clues when supervised with sufficient training data. Unlike the predominant image-text prediction methods~\cite{radford2021learning,zhou2022learning}, our model in~\figref{fig:comp} c) relies on the joint text-vision prototypes instead of the conventional $\bs{fc}$ layers. Considering the alignment during contrastive learning, we directly fuse the visual prototypes $\br{U}\in \mathbb{R}^{|\mc{C}|\times D_V}$ and $\br{T}\in \mathbb{R}^{|\mc{C}|\times D_T}$ with a learnable weight $\tau$. Hence for any query image $\br{x}$, the joint similarity scores from Eqn.~\eqref{eq:clip} are updated as:
\begin{equation}\label{eq:fusion}
\hat{\br{p}_{i}}=\frac{e^{\tau \cdot \xi(\mc{V}(\br{x}),\br{T}_{i}^{\top}) + \xi(\mc{V}(\br{x}),\br{U}_{i}^{\top})}}{\sum_{j=i}^{N} e^{\tau \cdot \xi(\mc{V}(\br{x}),\br{T}_{j}^{\top}) + \xi(\mc{V}(\br{x}),\br{U}_{j}^{\top})}}.
\end{equation}
This operation can be theoretically replaced by other concatenation or attention-based fusion strategies. Despite its simplicity, we found it works well under different scenarios, which are discussed later. We use these fused text-vision prototypes for both training and inference during base and incremental sessions. 
Benefiting from this prototypical design, the text knowledge can be taken as a part of visual encoders, and during inference time, we only use the visual backbones without any additional computation costs.

\begin{figure}
\begin{center}
\includegraphics[width=1\columnwidth]{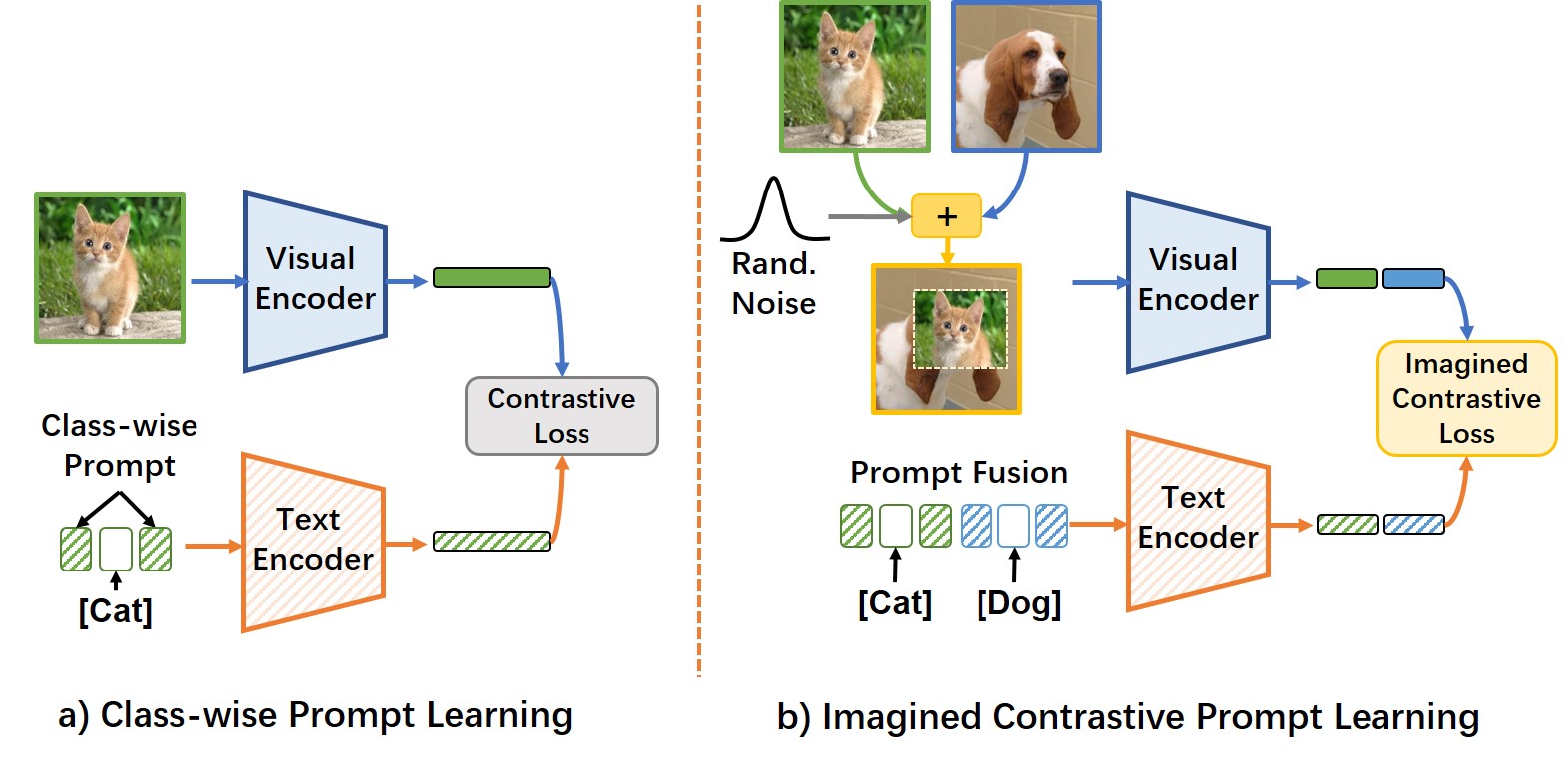}
 \caption{Illustrations of different text-vision learning loss. a) Class-wise context prompt Learning. b) Multi-modality imagined contrastive learning: two images using mixing strategy~\cite{yun2019cutmix} are aligned with their corresponding prompt fusion texts.
 }\label{fig:mix}
 \end{center}

\end{figure}

\subsection{Aligning Text with Image in FSCIL} \label{sec:align}
Vanilla contrastive learning adopts the handcrafted text prompt,~\eg, '$\bs{a photo of a [cls].}$' to get language embedding, which is aligned in the same space during the contrastive pretraining. However, during the downstream supervised learning process on visual encoders, it accompanies a clear domain gap between the text and vision embeddings. 
To solve this, we propose to find the generalized alignment strategy when only texts of labels (\eg, $[\bs{cat}]$) are available for training.

We make two major improvements for this multimodal alignment in FSCIL: 1) during the incremental learning session, we propose a context prompt learning method (\figref{fig:overview} a)) for fast adaptation of pretrained language knowledge on few-shot novel classes; 2) during the base training session, we propose the imagined contrastive learning (\figref{fig:overview} b)) to alleviate the imbalance of multi-modality data (\ie, sufficient visual training data while insufficient text descriptions). Besides, we also propose a space reservation in the base session to construct compact base prototypes to ``reserve" space for subsequent incremental prototypes.

\textbf{Imagined Contrastive Learning.} The other aforementioned challenge is caused by the insufficiency of text inputs compared to image data. The contrastive learning in~\figref{fig:mix} a) is easy to overfit on training data when only label text is available,~\eg, $N$ text phrases for $N$ classes. To alleviate this phenomenon, we introduce a contrastive loss conducted during imagined texts and images in~\figref{fig:mix} b), which is composed of two steps. i) the text prompts of two random classes are concatenated including the learnable ones:
\begin{equation}
f(\br{t}_i,\br{t}_j) = [\br{o}^{p}_{i}, \bs{cls}_{i} ,\br{o}^{s}_{i}, \br{o}^{p}_{j}, \bs{cls}_{j} ,\br{o}^{s}_{j}].
\end{equation}
ii) two visual images are fused averaged using CutMix~\cite{yun2019cutmix} or other alternative intra-mixing methods:
\begin{equation}
m(\br{x}_i,\br{x}_j) = \br{M}\cdot \br{x}_i +(\mathbbm 1-\br{M})\cdot \br{x}_j,
\end{equation} 
where $\br{M} \in \mathbb{R}^{H'\times W'}$ denotes the sampled masks. We control the mask proportions $\frac{H'\times W'}{H\times W} $ of $\br{M}$ are sampled from $(0.4, 0.6)$ to match the text concatenation while introducing randomness. For two mixed samples $i,j$ in batch $\mc{B}$, the imagined contrastive learning has the form:
\begin{equation}\label{eq:imagined}
 \mc{L}_{\text{im}}( \mc{B};\br{o},\theta) = -\sum_{i,j} \log \frac{e^{\xi(\mc{V}(m(\br{x}_i,\br{x}_j);\theta),\mc{T}(f(\br{t}_i,\br{t}_j;\br{o})))}}{\sum_{p,q\in \mc{B}} e^ {\xi(\mc{V}(m(\br{x}_i,\br{x}_j);\theta),\mc{T}(f(\br{t}_p,\br{t}_q;\br{o})))}}.
\end{equation}
While for each positive pair $m(\br{x}_{i},\br{x}_{j})$ and $f(\br{t}_{i},\br{t}_{j})$, the negative samples are other mixtures with different $p\ne i, q\ne j$ in the same batch.

\textbf{Learning with Space Reservation.} Besides the aforementioned contrastive learning, here we introduce the auxiliary margin-based cross-entropy  $\mc{L}_{\text{M-CE}}$ for space reservation for incremental classes, which helps to project the classes into a normalized hypersphere. Here we introduce the margin-based softmax loss,~\eg, ArcFace~\cite{deng2019arcface} to help the ``space reservation", which alleviates the overfitting on base classes. This indicates that the base classes would be distributed more compactly and would not fulfill the overall manifold space, thus the reserved space can be retained for representing the incremental classes.
To be specific, this margin loss has the following form:
\begin{equation}
    \mc{L}_{\text{M-CE}} = -\log \frac{e^{s\cos{(\alpha_{\br{y}_i}+m)}}}{e^{s\cos{(\alpha_{\br{y}_i}+m)}}+\sum_{j=1,j\ne \br{y}_i}^{N} e^{s\cos{(\alpha_{j})}}},
\end{equation}
where we empirically set the scale $s$ as 1 with $m=0.4$ to maintain the magnitude of loss constraints. The $\cos \alpha_{j} $ denotes the cosine similarity between prototype $\hat{\br{p}_{j}}$ and visual features $\br{U}_{j}$.

\textbf{Context Prompt for Fast Adaptation.} During incremental sessions in FSCIL, only $M \le 5$ samples can be used for training, which makes the visual fine-tuning process a major obstacle. To start from another view, as the joint prediction in Eqn.~\eqref{eq:fusion} is also determined by the text embeddings $\br{T}=\mc{T}(\textbf{t})$, we propose to construct class-wise learnable prompt instead of the handcrafted ones in CLIP. We empirically use prefix $\br{o}^{p}$ and suffix $\br{o}^{s}$ learnable prompt for each class, which is accomplished as a whole learnable sentence $f(\br{t}_{i}) = [\br{o}^{p}_{i}, \bs{cls}_{i} ,\br{o}^{s}_{i}]$.
As in~\figref{fig:comp} c), during the incremental session, we now fix the visual and text encoders and only learnable context prompts for each class are fine-tuned to minimize the language-vision domain gap:
\begin{equation}\label{eq:fewshot}
 \mc{L}_{\text{conp}}(\cdot;\br{o}_{\{p,s\}})=  \Sigma_{(\br{x},\br{y},\br{t})\in \mc{D}^{s}} \br{y}\log \mc{S}(\xi(\mc{V}(\br{x}),\mc{T}(f(\br{t};\br{o})))),
\end{equation}
where $\mc{S}$ denotes the $\bs{SoftMax}$ function with pre-learnt weight $\tau$. With the class-wise learnable prompt, few-shot samples can be depicted with learnable sentences, and the domain gap of unseen categories is fast mitigated.

\begin{algorithm}[!t]
\caption{Language-inspired Relation Transfer (LRT)}\label{alg:lrt}
\hspace*{0.02in} {\bf Input:}
Base session dataset $\mc{D}^{1}=\{(\br{x}_{i}^{1},\br{y}_{i}^{k},\br{t}_{i}^{s})\}_{i=1}^{|\mc{D}^{1}|}$. Incremental session dataset $\mc{D}^{1} \ldots \mc{D}^S$.\\
\hspace*{0.02in} {\bf Output:}
Visual encoder:$\mc{V}$, text prompts:$\{\br{o}_{i}\}_{i=1}^{|\mc{C}|}$
\begin{algorithmic}[1]
\State Initialize Visual $\mc{V}(\cdot)$ and Text encoder $\mc{T}(\cdot)$ with CLIP alignment.

\Comment{{\color{blue}\textit{Base Session Pretraining}}}
\State Random Init. text prompt $f(\br{t}_{i}) = [\br{o}^{p}_{i}, \bs{cls}_{i} ,\br{o}^{s}_{i}]$
\State Random Init. visual prototypes $\br{V}$
\For{ $\forall ((\br{x}_{i}^{1},\br{y}_{i}^{k},\br{t}_{i}^{1}))  \in \mc{D}^{1}$}
\State Extract text features: $\br{T}_i = \mc{T}(f(\br{t}_{i})), i=1 \ldots |\mc{C}^1|$
\State Calculate text relationship $\br{A}_{i,j}$ by Eqn. ({\color{red}4})
\State Update visual prototypes $\br{V}$ as $\br{U}$ in Eqn. ({\color{red}5})
\State Fuse text-vision knowledge by Eqn. ({\color{red}6}) to obtain $\hat{\br{p}}$
\State Optimize $\mc{L}_{\text{CE}}(\hat{\br{p}},\br{y})$ and $\mc{L}_{\text{M-CE}}(\hat{\br{p}},\br{y})$
\State Construct mixed image $\hat{\br{x}}$ by CutMix
\Statex \qquad $m(\br{x}_i,\br{x}_j) = \br{M}\cdot \br{x}_i +(\mathbbm{1}-\br{M})\cdot \br{x}_j$
\State Construct mixed text prompt
\Statex \qquad $f(\br{t}_i,\br{t}_j) = [\br{o}^{p}_{i}, \bs{cls}_{i} ,\br{o}^{s}_{i}, \br{o}^{p}_{j}, \bs{cls}_{j} ,\br{o}^{s}_{j}]$
\State Calculate imagined contrastive loss $\mc{L}^{(i,j)}_{\text{im}}(\cdot;\br{o},\theta)$ in Eqn. ({\color{red}10})
\State Conduct base session optimization $\mc{L}_{\text{base}}$ in Eqn. ({\color{red}12}) with fixed text encoder $\mc{T}$
\EndFor

\Comment{{\color{blue}\textit{Incremental Session Fast Adaptation}}}

\State Transfer other learned weights and prompts to incremental session.
\State Random Init. text prompt $f(\br{t}_{i}) = [\br{o}^{p}_{i}, \bs{cls}_{i} ,\br{o}^{s}_{i}]$ for new classes $\mc{C}^{s}$
\For{ $\forall ((\br{x}_{i}^{s},\br{y}_{i}^{k},\br{t}_{i}^{s}))  \in \mc{D}^{s}$}
\State Init. visual prototypes $\br{V}$ by Eqn. ({\color{red}2}) using $\br{x}$
\State Extract text features: $\br{T}_i = \mc{T}(f(\br{t}_{i})), i=1 \ldots |\mc{C}^s|$
\State Conduct incremental session text-vision fast alignment $\mc{L}_{\text{inc}}=\mc{L}_{\text{conp}}(\cdot;\br{o}_{\{p,s\}})$ in Eqn. ({\color{red}11}) 
\EndFor
\State \Return Optimized visual encoder $\mc{V}$, text prompts $\{\br{o}_{i}\}_{i=1}^{|\mc{C}|}$
\end{algorithmic}
\end{algorithm}

\textbf{Overall Training Scheme.} The overall training follows the few-shot class-incremental learning paradigm. 1) During the base session, the visual prototypes are randomly initialized as classifiers and the learning objective is joint optimization of three terms~\ie, the cross-entropy $\mc{L}_{\text{CE}}$ between the fused image-text prediction and ground truth label, the space reservation constraints $\mc{L}_{\text{M-CE}}$, and the imagined contrastive learning $\mc{L}_{\text{im}}$:
\begin{equation}\label{eq:joint}
 \mc{L}_{\text{base}} =\mc{L}_{\text{CE}}(\hat{\br{p}},\br{y})+\lambda_{\text{m}}\mc{L}_{\text{M-CE}}(\hat{\br{p}},\br{y})+ \lambda_{\text{im}} \mc{L}_{\text{im}}(\br{x},\br{t}).
\end{equation}
The relational knowledge transfer module in~\secref{sec:relation} is consistent during the base and incremental sessions. These three loss functions are jointly optimized during the base session to construct a generalized text-to-image feature transferring space.
2) While in the incremental learning session, we first construct vision prototypes following Eqn.~\eqref{eq:visualproto} and then finetune the $ \mc{L}_{\text{inc}}=\mc{L}_{\text{conp}}(\cdot;\br{o}_{\{p,s\}})$ to fast mitigate the domain gap among text and vision while representing the visual samples with learnable text prompts. Note that we only conduct the prompt learning $\mc{L}_{\text{conp}}$ during incremental sessions, alleviating the overfitting of multi-modal feature space caused by extreme few-shot samples.
The detailed training algorithm is shown in Alg.~\ref{alg:lrt}, which adopts different training strategies during the base training session and incremental training session. As there are only a few visual samples for training, we only conduct the fast adaptation with a few learnable text prompts of the current training classes, while fixing the prompts of previously seen sessions.

For inference time, we first use learned text prompts $\{\br{o}_{i}\}_{i=1}^{|\mc{C}|}$ to update text prototypes and thus drop the heavy text encoder for inference, keeping other modules including knowledge transfer module consistent with the training phase. The final prediction is measured by fused text-vision prototypes in Eqn.~\eqref{eq:fusion}.

\begin{table*}[!t]
\centering{
\caption{Classification accuracy on \textit{mini}ImageNet dataset for 5-way 5-shot incremental learning. ${*}$: Performances reported by~\cite{tao2020few}.  $\Delta_{\text{imp}}$: averaged relative improvements across all sessions compared to the Finetune baseline. }\label{table:mini}
\footnotesize
\setlength{\tabcolsep}{2.8mm}
\renewcommand{\arraystretch}{1}
\resizebox{1\textwidth}{!}{
\begin{tabular}{ccccccccccccc}
\toprule
\multirow{2}{*}{\textbf{Method}}&\multirow{2}{*}{\textbf{Pub. Year}}& \multicolumn{9}{c}{\textbf{Accuracy in each session $\uparrow$}}& \multirow{2}{*}{\textbf{Avg.}}& \multirow{2}{*}{\textbf{$\Delta_{\text{imp}}$}}\\
\cline{3-11}
&&1 &2&3&4&5&6&7&8&9&  \\\hline
Finetune~\cite{tao2020few}&-&61.31& 27.22& 16.37& 6.08& 2.54& 1.56& 1.93& 2.60& 1.40& 13.45&$\textcolor{gray}{(+0.00)}$      \\
\hline
iCaRL~\cite{rebuffi2017icarl}$^{*}$&{\color{blue}CVPR 17}&  61.31&  46.32&  42.94&  37.63&  30.49&  24.00&  20.89&  18.80&  17.21&  33.28&$\textcolor{red}{(+19.84)}$ \\
Rebalance~\cite{hou2019learning} $^{*}$   &{\color{blue}CVPR 19}&  61.31& 47.80& 39.31& 31.91& 25.68 &21.35 &18.67 &17.24 &14.17 &30.83&$\textcolor{red}{(+17.38)}$ \\
TOPIC~\cite{tao2020few}&{\color{blue}CVPR 20}&  61.31& 50.09& 45.17& 41.16& 37.48& 35.52& 32.19& 29.46& 24.42& 39.64&$\textcolor{red}{(+26.19)}$  \\
FSLL+SS~\cite{mazumder2021few}&{\color{blue}CVPR 20}& 68.85& 63.14& 59.24 &55.23 &52.24& 49.65& 47.74 &45.23& 43.92& 53.92&$\textcolor{red}{(+40.47)}$ \\
IDLVQ-C~\cite{chen2020incremental}&{\color{blue}ICLR 20}& 64.77& 59.87& 55.93& 52.62& 49.88 &47.55 &44.83& 43.14& 41.84& 51.16&$\textcolor{red}{(+37.71)}$ \\
CEC~\cite{zhang2021few}&{\color{blue}CVPR 21}& 72.00 & 66.83&  62.97&  59.43 & 56.70&  53.73 & 51.19&  49.24 & 47.63 & 57.75&$\textcolor{red}{(+44.30)}$ \\
F2M~\cite{shi2021overcoming} &{\color{blue}NeurIPS 21}& 67.28&  63.80 & 60.38&  57.06 & 54.08&  51.39&  48.82&  46.58&  44.65&54.89 &$\textcolor{red}{(+41.44)}$ \\
MetaFSCIL~\cite{chi2022metafscil} &{\color{blue}CVPR 22}& 72.04& 67.94 & 63.77& 60.29 & 57.58& 55.16& 52.90&  50.79& 49.19&58.85& $\textcolor{red}{(+45.40)}$ \\
LIMIT~\cite{zhou2022few} &{\color{blue}TPAMI 22}& 72.32 &68.47 &64.30& 60.78& 57.95& 55.07& 52.70 &50.72& 49.19&59.05&$\textcolor{red}{(+45.61)}$ \\
FACT~\cite{zhou2022forward} &{\color{blue}CVPR 22}& 72.56& 69.63 & 66.38& 62.77 &60.60& 57.33& 54.34& 52.16& 50.49&60.70& $\textcolor{red}{(+47.25)}$ \\
C-FSCIL~\cite{hersche2022constrained} &{\color{blue}CVPR 22}& 76.40& 71.14 &66.46& 63.29 &60.42& 57.46& 54.78& 53.11& 51.41&61.61 &$\textcolor{red}{(+48.16)}$ \\

\midrule
Base-V (FT) &-& 72.88& 67.65& 63.09& 59.09& 55.54& 52.79 & 49.97& 47.87& 45.59&  57.16 &$\textcolor{red}{(+43.71)}$ \\
CLIP (0-shot) &-& 65.18& 65.05& 63.20& 62.58&62.49 &62.54 & 61.33 & 60.98& 60.62& 62.67&$\textcolor{red}{(+49.21)}$ \\
Ours (LRT) &-& \textbf{90.17}& \textbf{85.82} &\textbf{81.70}& \textbf{78.12}& \textbf{75.04}& \textbf{71.71}&\textbf{68.88}& \textbf{66.74}& \textbf{65.34}&\textbf{75.94}&$\textcolor{red}{\br{(+62.49)}}$ \\
\bottomrule
\end{tabular}
}
}
\end{table*}

\begin{table*}[!t]
\centering{
\caption{Classification accuracy on CIFAR100 dataset for 5-way 5-shot incremental learning. ${*}$: Performances are reported by~\cite{tao2020few}.  $\Delta_{\text{imp}}$: averaged relative improvements across all sessions compared to the Finetune baseline.}\label{table:mini}
\footnotesize
\renewcommand{\arraystretch}{1}
\resizebox{1\textwidth}{!}{
\begin{tabular}{ccccccccccccc}
\toprule
\multirow{2}{*}{\textbf{Method}}&\multirow{2}{*}{\textbf{Pub. Year}}& \multicolumn{9}{c}{\textbf{Accuracy in each session $\uparrow$}}& \multirow{2}{*}{\textbf{Avg.}}& \multirow{2}{*}{\textbf{$\Delta_{\text{imp}}$}}\\
\cline{3-11}
&&1 &2&3&4&5&6&7&8&9&  \\\hline
Finetune~\cite{tao2020few}&-&64.10& 39.61& 15.37& 9.80& 6.67& 3.80& 3.70& 3.14& 2.65& 16.53&$\textcolor{gray}{(+0.00)}$ \\
\hline
iCaRL~\cite{rebuffi2017icarl}$^{*}$&{\color{blue}CVPR 17}&  64.10 &53.28 &41.69& 34.13& 27.93 &25.06 &20.41& 15.48& 13.73 &32.87&$\textcolor{red}{(+16.33)}$ \\
Rebalance~\cite{hou2019learning} $^{*}$   &{\color{blue}CVPR 19}&  64.10 &53.05& 43.96 &36.97 &31.61& 26.73 &21.23 &16.78& 13.54 &34.21&$\textcolor{red}{(+17.68)}$ \\
TOPIC~\cite{tao2020few}&{\color{blue}CVPR 20}&  64.10& 55.88& 47.07& 45.16& 40.11& 36.38& 33.96& 31.55& 29.37&42.62&$\textcolor{red}{(+26.08)}$ \\
FSLL+SS~\cite{mazumder2021few}&{\color{blue}CVPR 20}& 66.76& 55.52& 52.20 &49.17 &46.23& 44.64& 43.07 &41.20& 39.57&48.71&$\textcolor{red}{(+32.17)}$\\
CEC~\cite{zhang2021few}&{\color{blue}CVPR 21}& 73.07& 68.88 &65.26& 61.19& 58.09& 55.57 &53.22 &51.34 &49.14&59.53&$\textcolor{red}{(+42.99)}$\\
F2M~\cite{shi2021overcoming} &{\color{blue}NeurIPS 21}& 61.71&  62.05 & 59.01&  55.58 & 52.55&  49.96&  48.08&  46.28&  44.67&53.32&$\textcolor{red}{(+36.78)}$\\
DSN~\cite{yang2022dynamic} &{\color{blue}TPAMI 22}& 73.00& 68.83& 64.82& 62.24& 59.16& 56.96& 54.04& 51.57& 49.35&60.00 &$\textcolor{red}{(+43.47)}$ \\
MetaFSCIL~\cite{chi2022metafscil} &{\color{blue}CVPR 22}& 74.50& 70.10 & 66.84&  62.77 & 59.48&  56.52&  54.36&  52.56&  49.97&60.79&$\textcolor{red}{(+44.25)}$\\
C-FSCIL~\cite{hersche2022constrained} &{\color{blue}CVPR 22}& 77.47& 72.40 &67.47& 63.25 &59.84& 56.95& 54.42& 52.47& 50.47&61.64&$\textcolor{red}{(+45.10)}$ \\
LIMIT~\cite{zhou2022few} &{\color{blue}TPAMI 22}& 73.81& 72.09& 67.87& 63.89& 60.70 &57.78& 55.68& 53.56 &51.23&61.84 &$\textcolor{red}{(+45.31)}$ \\
FACT~\cite{zhou2022forward} &{\color{blue}CVPR 22}& 74.60& 72.09 &67.56& 63.52 &61.38& 58.36& 56.28& 54.24& 52.10&62.23&$\textcolor{red}{(+45.70)}$ \\

\midrule
Base-V (FT) &-&67.37& 62.37& 58.00& 54.27& 51.11& 48.32& 45.71& 43.57& 41.50&52.47&$\textcolor{red}{(+35.93)}$\\
CLIP (0-shot) &-& 39.78&38.25&36.97&34.32&33.26&32.07&31.99&31.24&30.24&34.24&$\textcolor{red}{(+17.71)}$ \\
Ours (LRT) &-&\textbf{87.02}& \textbf{82.40}& \textbf{77.84}& \textbf{73.31}& \textbf{70.18}& \textbf{66.74}& \textbf{64.50}& \textbf{61.99}& \textbf{59.49}&\textbf{71.50}&$\textcolor{red}{\br{(+54.96)}}$\\
\bottomrule
\end{tabular}
}}
\end{table*}

\section{Experiments}\label{sec:exp}
\subsection{Experimental settings}

\textbf{Dataset and Evaluations.} In this experiment, following the splits in prevailing works~\cite{tao2020few,hersche2022constrained,zhang2021few}, we mainly conduct ablations on two widely-used benchmark datasets,~\ie, \textit{mini}ImageNet~\cite{deng2009imagenet} and CIFAR-100 dataset~\cite{krizhevsky2009learning}. \textit{mini}ImageNet~\cite{deng2009imagenet} contains 100 different semantic classes, which are divided into 60 base classes and 40 few-shot classes for 8 incremental sessions. In the base sessions, each class has 600 images with a resolution of $84\times 84$, while in the few-shot session, only 5 images of each class are used for training. Besides, we conduct experiments on the large-scale ImageNet100~\cite{deng2009imagenet} dataset of over 128k images following~\cite{zhou2022forward} with the image resolution of $224\times 224$. Similarly, ImageNet100 contains 100 different semantic classes, which are divided into 60 base classes and 40 few-shot classes for 8 incremental sessions. Besides, the CIFAR-100 dataset~\cite{krizhevsky2009learning} is also divided into 60 base classes and 40 few-shot incremental classes, with the resolution of $32\times 32$.
The final evaluations are conducted on classes across all training sessions.

\textbf{Implementation Details.} To conduct fair comparisons with state-of-the-art works, we follow~\cite{zhang2021few} to conduct the supervised training with identical data augmentation strategies.
We adopt the lightweight ResNet-50~\cite{he2016deep} model pretrained by CLIP~\cite{radford2021learning} to alleviate the additional parameters. The text prompt is set as '$\bs{a photo of a [cls].}$' for fair comparisons with prevailing works. Following~\cite{zhang2021few}, The model is trained with the batch size of 128 with SGD for 100 epochs. The learning rate starts at 0.01 for both \textit{mini}ImageNet~\cite{deng2009imagenet} and CIFAR-100 dataset~\cite{krizhevsky2009learning} and decays at 40 and 70 epochs. The text encoders are fixed across all the sessions, and the learnable prompt length is set as 4. Balanced weights $\lambda_{m}$, $\lambda_{im}$ are set as 0.1 and 0.05 respectively. We resize the low-resolution image ($84\times 84$) to fit the positional encoding layer of CLIP models.

\subsection{Comparison with State-of-the-art}

\textbf{Results on \textit{mini}ImageNet.} In~\tabref{table:mini}, we first conduct experiments on the widely-used challenging \textit{mini}ImageNet dataset with state-of-the-art works, including several CIL methods~\cite{rebuffi2017icarl,hou2019learning} and FSCIL methods~\cite{tao2020few,mazumder2021few,zhang2021few,chi2022metafscil,hersche2022constrained}. Pioneer works~\cite{tao2020few} indicate learning with a naive finetuning strategy in the first line would lead to catastrophic forgetting on base sessions, while the CIL methods alleviate this difficulty by clear improvements. To validate the effectiveness of our method, we conduct a baseline finetuning only using the visual encoders (ResNet-CLIP) using the identical protocol (\textbf{Base-V}) with prototypical learning of~\figref{fig:comp} a) in the incremental session, which shows slightly higher performance than earlier models. The zero-shot CLIP in~\figref{fig:comp} b)  have a strong generalization ability and do not need any training data. The last session's accuracy of zero-shot CLIP remains at high accuracy.
With our proposed Language-inspired Relation Transfer (LRT) model, the performance of the last session is improved by $19.7\%$, and also shows a clear margin~\ie, $13.9\%$ and $14.8\%$ compared to the state-of-the-art C-FSCIL~\cite{hersche2022constrained} and FACT~\cite{zhou2022forward} methods.

\begin{table*}[!t]
\centering{
\caption{Comparisons on ImageNet100 dataset for 5-way 5-shot incremental learning. *: Methods are implemented using official codes.   }\label{table:imagenet100}
\footnotesize
\setlength{\tabcolsep}{3.8mm}
\renewcommand{\arraystretch}{1.0}
\resizebox{1\textwidth}{!}{
\begin{tabular}{ccccccccccc}
\toprule
\multirow{2}{*}{\textbf{Method}}& \multicolumn{9}{c}{\textbf{Accuracy in ImageNet100 $\uparrow$}}& \multirow{2}{*}{\textbf{Avg.}}\\
\cline{2-10}
&1 &2&3&4&5&6&7&8&9&\\\hline
CEC~\cite{zhang2021few}*&84.77& 80.03& 76.66& 73.10&69.30& 65.88& 64.27& 62.91& 60.04 &  70.77 \\
FACT~\cite{zhou2022forward}* & 86.00& 80.94 & 77.66& 75.34 &70.40& 66.72& 64.82& 63.15& 60.98&71.67 \\
Ours (LRT)& \textbf{91.43}& \textbf{87.03} &\textbf{83.83}& \textbf{79.34}& \textbf{76.15}& \textbf{72.05}&\textbf{70.18}& \textbf{68.52}& \textbf{65.90}&\textbf{77.16}\\
\bottomrule
\end{tabular}
}
}
\end{table*}

\begin{figure}[!t]
\begin{center}
\includegraphics[width=1\columnwidth]{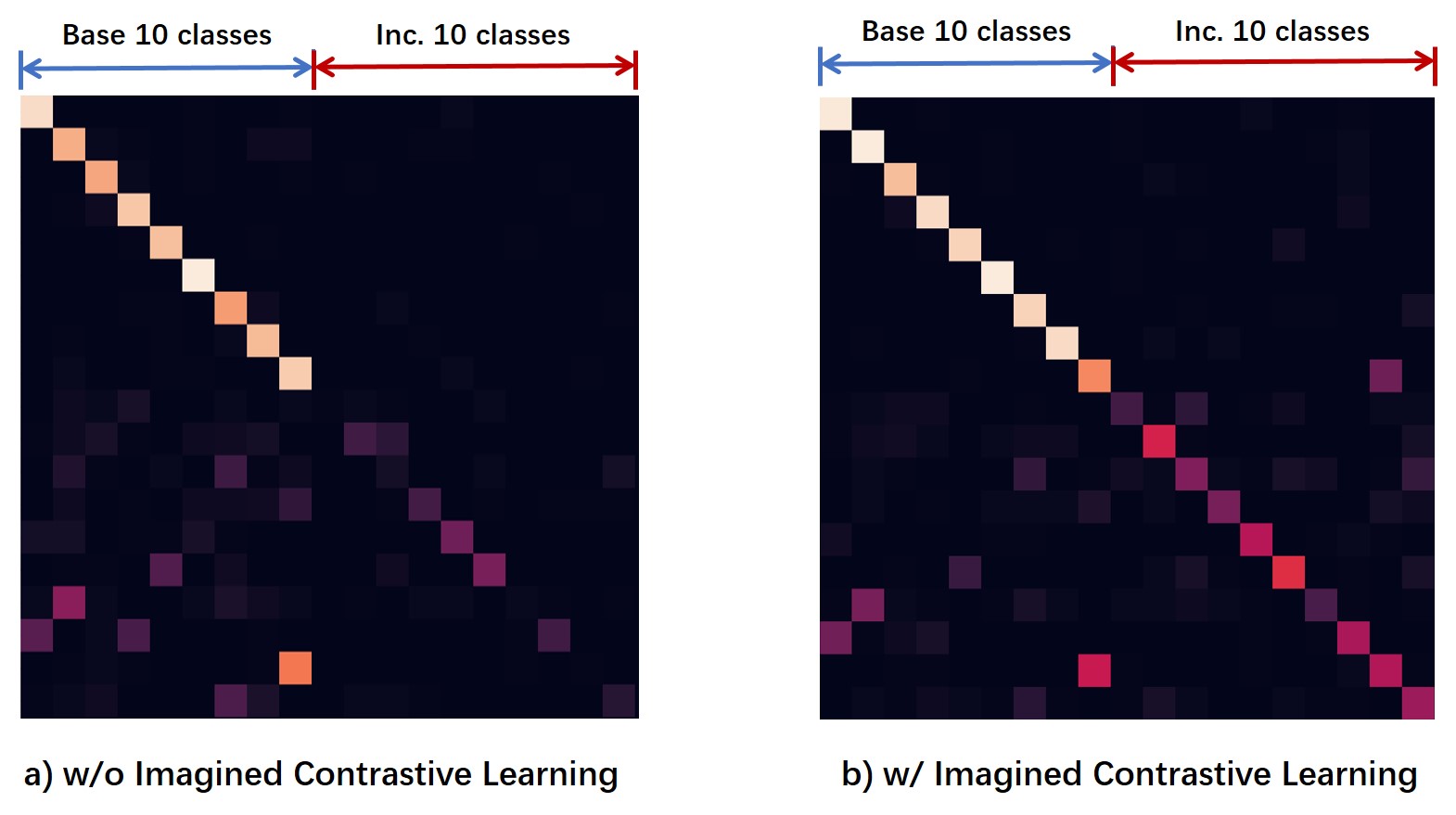}
 \caption{Confusion matrices of different contrastive losses. The last 10 base classes and the first 10 incremental classes on \textit{mini}ImageNet are zoomed in for ablation comparisons.
 }\label{fig:exp_mix}
 \end{center}
\end{figure}

\textbf{Results on CIFAR100.} Note that CIFAR-100 is a low-resolution image recognition dataset and CLIP models face difficulties in conducting zero-shot testing. The first session shows about a $25\%$ gap in accuracy and Base-V with fine-tuning surpasses the zero-shot CLIP models by $18\%$ in the average accuracy. Similar to the performances on \textit{mini}ImageNet, our model also shows a notable improvement on the CIFAR100 dataset. Existing models including CEC~\cite{zhang2021few}, FACT~\cite{zhou2022forward} with fixed or slightly tuned encoders (\figref{fig:comp} a)) shows a clear performance improvement compared to the topological re-adjusting ones~\cite{tao2020few}. As this prevailing trend shows a performance bottleneck, our LRT steadily improves the performances of baseline finetuning (\textbf{Base-V}) by nearly $18.0\%$ and surpasses the second best method~\cite{zhou2022forward} by $7.3\%$ in the last session.

\textbf{Results on ImageNet100.} To verify the performance on large-scale datasets, we compare our proposed method with the source code of state-of-the-art methods CEC~\cite{zhang2021few}, FACT~\cite{zhou2022forward} and adopt the same augmentation from~\cite{zhou2022forward}. As in~\tabref{table:imagenet100} FACT~\cite{zhou2022forward} improves the base and incremental learning sessions by $0.9\%$, while our proposed method (LRT) shows steady improvements and surpasses FACT~\cite{zhou2022forward} by a clear margin of $5.49\%$, implying the good potential of our model for large-scale datasets.

\begin{table*}[t]
\centering{
\caption{Ablation studies on \textit{mini}ImageNet and CIFAR100 benchmarks. PT: learnable context prompt finetuning. FT: standard visual finetuning. $\mc{M}_{fuse}$, $\mc{M}_{graph}$: the text-vision prototypical fusion and graph relationship transformation. Acc $(\mc{D}^S)$: accuracy of the last session. Avg. Acc: averaged accuracy of all sessions. $\Delta_{avg}$: relative improvements.}
\label{table:ablation}
\setlength{\tabcolsep}{2.0mm}
\renewcommand{\arraystretch}{1.0}
\footnotesize
\resizebox{1\textwidth}{!}{
\begin{tabular}{cccc|ccc|ccc}
\toprule
\multirow{2}{*}{$\bs{Text}$}&\multirow{2}{*}{$\bs{Vision}$} &\multirow{2}{*}{$\bs{T-V Trans}$}& \multirow{2}{*}{$\bs{T-V Loss}$}& \multicolumn{3}{c}{\textit{mini}ImageNet}& \multicolumn{3}{c}{CIFAR100} \\
&&& &  Acc $(\mc{D}^S)$ & Avg. Acc& $\Delta_{avg}$&  Acc $(\mc{D}^S)$ & Avg. Acc& $\Delta_{avg}$ \\
\midrule
-&FT&Only V&$\mc{L}_{\text{CE}}(\bs{V})$&45.59&57.16&$\textcolor{gray}{(+0.00)}$&41.50&52.47&$\textcolor{gray}{(+0.00)}$\\
PT&Fixed&Only T&
$\mc{L}_{\text{CE}}$(Eqn.\eqref{eq:clip})&42.54&52.87&$\textcolor{blue}{(-4.30)}$&38.83&49.86&$\textcolor{blue}{(-2.61)}$\\
Fixed&FT&$\mc{M}_{fuse}$& 
$\mc{L}_{\text{CE}}$(Eqn.\eqref{eq:fusion})
&50.60&64.98&$\textcolor{red}{(+7.81)}$&50.74&64.77&$\textcolor{red}{(+12.31)}$\\
PT&FT&$\mc{M}_{fuse}$&
$\mc{L}_{\text{CE}}$(Eqn.\eqref{eq:fusion})&54.33&67.01&$\textcolor{red}{(+9.85)}$&54.69&67.32&$\textcolor{red}{(+14.84)}$\\
PT&FT&$\mc{M}_{fuse}$&Eqn.\eqref{eq:joint}&56.00&67.88&$\textcolor{red}{(+10.72)}$&55.18&67.67&$\textcolor{red}{(+15.20)}$\\
PT&FT&$\mc{M}_{fuse}$+$\mc{M}_{graph}$&Eqn.\eqref{eq:joint}&\textbf{65.34}&\textbf{75.94}&$\textcolor{red}{(\textbf{+18.78})}$&\textbf{59.49}&\textbf{71.50}&$\textcolor{red}{(\textbf{+19.03})}$\\
\bottomrule
\end{tabular}
}
}
\end{table*}

\begin{table}[t]
\centering{
\caption{ Performance analysis of different Text-vision ($\bs{T}\rightarrow \bs{I}$) methods and Prototypical fusion strategies on \textit{mini}ImageNet. Proto Add. : averaged summation of vision and text prototypes. Static: Learnable $\tau$ in Eqn.~\eqref{eq:fusion} is set as 1.}
\label{table:t2i}
\setlength{\tabcolsep}{0.7mm}
\renewcommand{\arraystretch}{1.0}
\footnotesize
\resizebox{1\linewidth}{!}{
\begin{tabular}{cc|c|c|c}
\toprule
{$\bs{T}\rightarrow \bs{I}$ Methods} & Strategy&  Acc $(\mc{D}^S)$ & Avg. Acc& $\Delta_{avg}$\\
\midrule
Baseline (w/o $\bs{T}$)& - &45.59&57.16&$\textcolor{gray}{(+0.00)}$\\
$\mc{M}_{graph}+\mc{M}_{fuse}$& Proto Add.&42.74&51.25&$\textcolor{blue}{(-5.91)}$\\
$\mc{M}_{graph}+\mc{M}_{fuse}$& Static&45.46&51.10&$\textcolor{blue}{(-6.06)}$\\
$\mc{M}_{fuse}$ & Learnable&56.00&67.88&$\textcolor{red}{(+10.72)}$\\
$\mc{M}_{graph}$& Learnable&60.30&72.61&$\textcolor{red}{(+15.45)}$\\
$\mc{M}_{graph}+\mc{M}_{fuse}$& Learnable&65.34&75.94&$\textcolor{red}{(+18.78)}$\\
\bottomrule
\end{tabular}
}
}
\end{table}

\begin{figure}[!t]
\begin{center}
\includegraphics[width=0.93\columnwidth]{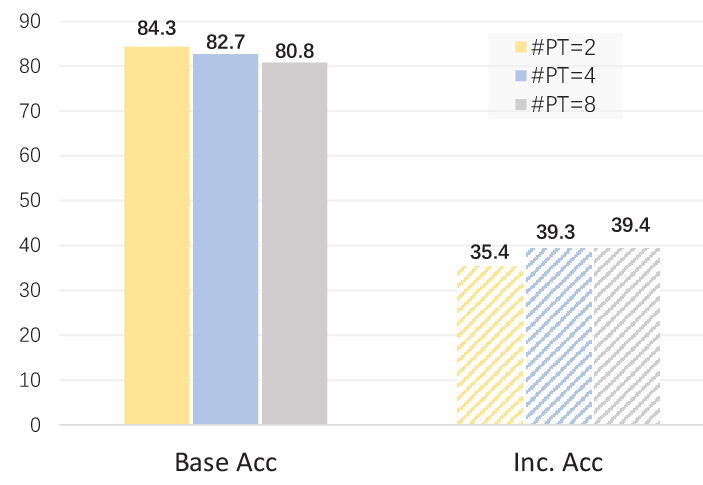}
 \caption{Accuracies of base and incremental sessions with different learnable prompt lengths. Our method chooses prompt length $\br{\#\text{PT}}=4$ for the base and incremental trade-off.
 }\label{fig:length}
 \end{center}
\end{figure}

\subsection{Performance Analysis}

\textbf{Ablations of Learning Paradigms.} In~\tabref{table:ablation}, we conduct detailed ablations of our proposed learning paradigms. The first two lines show methods only using visual FineTuning (FT) and text Prompt Tuning (PT) respectively, which show inferior results on both base and incremental sessions. With the addition of the knowledge fusion module in the third line, the model shows a clear performance improvement,~\eg, over $9.0\%$ on the last session of CIFAR compared to the Base-V. The fourth and fifth line introduces the joint PT-visual tuning and the imagined contrastive loss, which presents steady improvements on both last session accuracy and average accuracy.
The final line is our full model which can still boost the performance by using graph transformations (over $18\%$ than visual baselines).

\textbf{Effects of Text-Vision Relationship.} We first conduct a visualized ablations with the Imagined Contrastive Training Loss of Eqn.~\eqref{eq:joint}. In~\figref{fig:exp_mix} a), we exhibit results that adopt the standard cross entropy $\mc{L}_{\text{CE}}(\bs{V})$ for only using vision prototypes, and Eqns.~\eqref{eq:fusion} and~\eqref{eq:joint} denote the only image-text fusion and fusion with imagined contrastive loss. The upper left diagonal shows the last 10 base classes in the miniImageNet dataset, and the lower shows the confusion matrices on 10 new classes. Comparing~\figref{fig:exp_mix} a) with b), it can be found our proposed alignment strategies greatly improve the learning of new classes without forgetting the base knowledge.

Besides visualization of loss functions, the fusion strategies are also important in our method, here we present several naive implementations in~\tabref{table:t2i}. The first line shows the baseline visual fine-tuning model without the help of text information. The \textbf{Learnable} denotes our fusion methods using Eqn.~\eqref{eq:fusion} and the \textbf{Static} denotes a fixed $\tau=1$. The \textbf{Proto Add.} denotes we directly add two prototypes before measurement,~\ie, $\xi(\br{x},\br{V}+\br{T})$. With the proposed graph module, the overall accuracy of base and incremental sessions shows a clear improvement. 
The experimental evidence shows the performance drops dramatically without the proper fusion of text and vision embedding since they are strongly aligned during the pretraining stage.

\textbf{Effects of Context Prompt.} Selecting the proper size of the prompt length is one of the factors that affect the final performance. We conduct hyper-parameter ablations by setting the prompt length (both prefix and suffix, $\textbf{\text{\#PT}}$) as 2,4 and 8 respectively. ~\figref{fig:length} shows that using more learnable prompts can boost the incremental learning session with only a few given samples,~\eg, $39.3\%$ vs. $35.4\%$. Nevertheless, finetuning more prompts will lead to the loss of base knowledge,~\eg, $1.6\%$ when extending the prompt number from 2 to 4. To achieve a good trade-off between the base and incremental sessions, we chose the prompt length of 4 in all our experiments.

\begin{table}[t]
\centering{
\caption{Performance analysis of different incremental shots on \textit{mini}ImageNet. $\Delta_{avg}$: relative improvements of averaged accuracy. }
\label{table:shots}
\setlength{\tabcolsep}{0.7mm}
\renewcommand{\arraystretch}{1.0}
\footnotesize
\resizebox{1\linewidth}{!}{
\begin{tabular}{c|c|c|cc}
\toprule
Incremental Shots&  Base. Acc & Inc. Acc& Avg. Acc& $\Delta_{avg}$\\
\midrule
5-shot&82.68&\textbf{39.32}&\textbf{61.00}&$\textcolor{gray}{(+0.00)}$\\
4-shot&83.15&37.42&60.28&$\textcolor{blue}{(-0.72)}$\\
3-shot&83.77&34.02&58.90&$\textcolor{blue}{(-2.10)}$\\
2-shot&84.22&27.40&55.81&$\textcolor{blue}{(-5.19)}$\\
1-shot&\textbf{86.40}&17.12&51.76&$\textcolor{blue}{(-9.24)}$\\
\bottomrule
\end{tabular}
}}
\end{table}

\begin{figure}[t]
\begin{center}
\includegraphics[width=0.85\columnwidth]{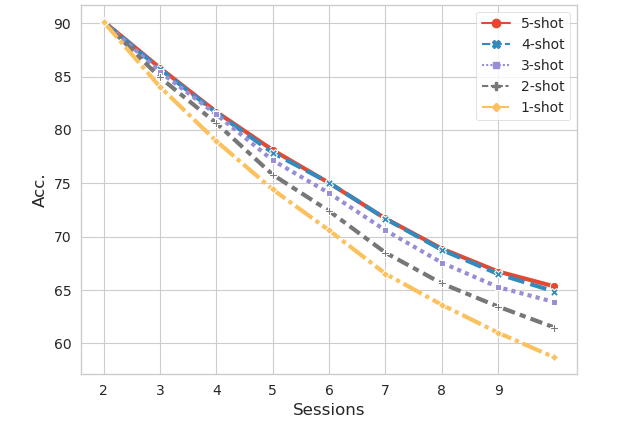}
 \caption{Accuracy with different number of shots during incremental sessions on \textit{mini}ImageNet dataset.  
 }\label{fig:shots}
 \end{center}

\end{figure}

\textbf{Incremental Learning with Fewer Shots.}
Besides the exploration of common N-way 5-shot settings during incremental learning. Here we exhibit the results on fewer shots in~\figref{fig:shots},~\ie, from 1-shot to 4-shots. Starting from the same base accuracy of nearly $90\%$, the model with fewer shots performs a more notable performance drop than ours with 5 training shots. Whilst it is still acceptable (over $51\%$ in Avg. Acc) compared to other earlier methods.
However, when considering the base accuracy and incremental accuracy individually in~\tabref{table:shots}, the performance of incremental sessions drops significantly. With the decrease of training shots, the accuracy on base sessions seems to retain the pretrained representations in base sessions, which shows higher performance ($86.40\%$ of 1-shot compared to $82.68\%$ of 5-shots). But the accuracy of incremental sessions drops dramatically. This indicates that it is hard to align the visual concepts with proper text descriptions with only one sample. In other words, depicting objects with pretrained textual knowledge also needs more visual cases to alleviate overfitting.

\begin{table*}[t]
\centering{
\caption{Knowledge transfer verification on \textit{mini}ImageNet benchmarks. PT: learnable context prompt finetuning. FT: standard visual finetuning. $\mc{M}_{fuse}$, $\mc{M}_{graph}$: the text-vision prototypical fusion and graph relationship transformation.  $\Delta_{har}$: relative improvements of harmonic accuracy.}
\label{tab:trans-ablation}
\setlength{\tabcolsep}{3.5mm}
\renewcommand{\arraystretch}{1.1}
\footnotesize
\resizebox{0.9\textwidth}{!}{
\begin{tabular}{cccc|cccc}
\toprule
\multirow{2}{*}{$\bs{Text}$}&\multirow{2}{*}{$\bs{Vision}$} &\multirow{2}{*}{$\bs{T-V Trans}$}& \multirow{2}{*}{$\bs{T-V Loss}$}& \multicolumn{4}{c}{\textit{mini}ImageNet} \\
&&& &  Base Acc. & Inc. Acc.& Har. Acc.& $\Delta_{har}$ \\
\midrule
PT&FT&$\mc{M}_{fuse}$&Eqn.\eqref{eq:fusion}&\textbf{83.78}&10.15&18.11&$\textcolor{gray}{(+0.00)}$\\
PT&FT&$\mc{M}_{fuse}$&Eqn.\eqref{eq:joint}&76.97&24.55&37.07&$\textcolor{red}{(+18.96)}$\\
PT&FT&$\mc{M}_{fuse}$+$\mc{M}_{graph}$&Eqn.\eqref{eq:joint}&82.68&\textbf{39.32}&\textbf{53.29}&$\textcolor{red}{(+35.18)}$\\
\bottomrule
\end{tabular}
}
}
\end{table*}

\begin{figure}[!t]
	\centering
	\includegraphics[width=.85\columnwidth]{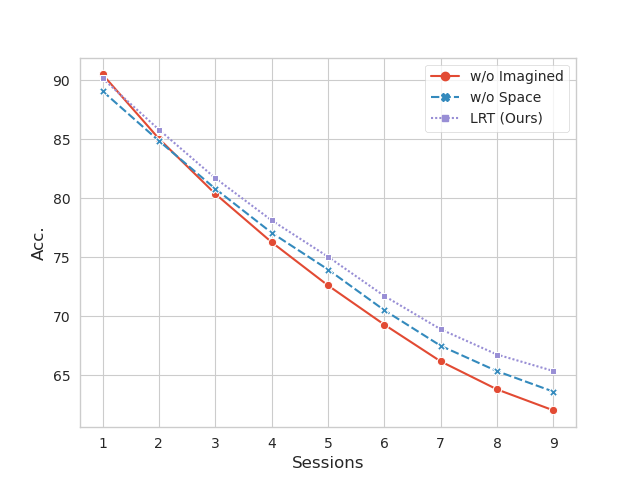}
	\caption{Ablations of different loss functions. Imagined: imagined contrastive learning $\mc{L}_{\text{im}}$. Space: space reservation loss $\mc{L}_{\text{M-CE}}$.
	}
	\label{fig:eq12}
\end{figure}

\textbf{Effect of Loss Constraints.} we conduct experiments on different gratitude of hyperparameters~\ie,  $\lambda_{\text{im}}$ for $\mc{L}_{\text{im}}$ and $\lambda_{\text{m}}$ for $\mc{L}_{\text{M-CE}}$.
The experimental results can be found in~\tabref{table:hyper}. Enlarging or reducing the balanced weight would lead to a clear performance drop of 1.8\% to 4.05\%. Besides, only scaling up the $\mc{L}_{\text{M-CE}}$ constraints would lead to a clear base session drop, while other imagined contrastive learning and space reservation constraints mainly affect the ability to learn new concepts. The detailed ablations of $\mc{L}_{\text{im}}$ and $\mc{L}_{\text{M-CE}}$ in Eqn.~\eqref{eq:joint} is presented in~\figref{fig:eq12}. With the collaborative learning of these loss functions, the capability to learn novel concepts has been greatly enhanced.

\begin{table}[!t]
\centering{
\caption{Performance analysis of balanced weights $\lambda_\text{m}$ and $\lambda_\text{im}$ in Eqn.~\eqref{eq:joint}.}
\label{table:hyper}
\setlength{\tabcolsep}{1.3mm}
\renewcommand{\arraystretch}{1.0}
\footnotesize
\resizebox{1\linewidth}{!}{
\begin{tabular}{ll|c|c|c|c}
\toprule
 $\lambda_{\text{im}}$ &  $\lambda_{\text{m}}$&  Acc $(\mc{D}^1)$ &  Acc $(\mc{D}^S)$ & Avg. Acc& $\Delta_{avg}$\\
\midrule
 $0.1\times$& $1\times$&89.93&61.28&73.49&$\textcolor{blue}{(-2.45)}$\\
$1\times$ &$0.1\times$&90.05&62.74&74.13&$\textcolor{blue}{(-1.81)}$\\
$1\times$& $10\times$&87.63&61.13&71.89&$\textcolor{blue}{(-4.05)}$\\
$10\times$& $1\times$&90.02&62.27&73.78&$\textcolor{blue}{(-2.16)}$\\
$1\times$& $1\times$&90.17&65.34&75.94&$\textcolor{gray}{(+0.0)}$\\
\bottomrule
\end{tabular}
}}
\end{table}

\begin{table}[!t]
\centering{
\caption{Different measurement comparisons on public benchmarks. Models are evaluated using public codes.}
\label{tab:har}
\setlength{\tabcolsep}{1.1mm}
\renewcommand{\arraystretch}{1.0}
\footnotesize
\resizebox{1\linewidth}{!}{
\begin{tabular}{c|c|c|c|c|c|c}
\toprule
\textbf{Dataset}&\textbf{Method}&  $(\mc{D}^1)$ &  $(\mc{D}^S)$ & Base Acc.& Inc Acc. & Har. Acc.\\
\midrule
\multirow{4}{*}{mini-IN}&CEC~\cite{zhang2021few}  &72.25&47.67&67.97&17.23&27.49\\
&FACT~\cite{zhou2022forward} &75.23&48.61&72.62&12.60&21.47\\
&Ours-2PT&89.95&64.74&\textbf{84.28}&35.42&49.88\\
&Ours-4PT&\textbf{90.17}&\textbf{65.34}&82.68&\textbf{39.32}&\textbf{53.29}\\
\midrule
\multirow{3}{*}{CIFAR}&CEC~\cite{zhang2021few}   &73.07&49.10&67.90&20.90&31.96\\
&FACT~\cite{zhou2022forward}&78.65&51.19&71.02&21.45&32.94\\
&Ours-4PT&\textbf{87.02}&\textbf{59.49}&\textbf{78.68}&\textbf{30.70}&\textbf{44.17}\\
\bottomrule
\end{tabular}
}}
\end{table}

\begin{table*}[!t]
\centering{
\caption{Comparisons of standard training and finetuning with CLIP visual backbones on \textit{mini}ImageNet dataset for 5-way 5-shot incremental learning. Standard Base: visual learning baseline proposed in~\secref{sec:pre}.}\label{table:base}
\footnotesize
\renewcommand{\arraystretch}{1}
\resizebox{1\textwidth}{!}{
\begin{tabular}{cccccccccccc}
\toprule
\multirow{2}{*}{\textbf{Method}}&\multirow{2}{*}{\textbf{Backbone}}& \multicolumn{9}{c}{\textbf{Accuracy in mini-ImageNet $\uparrow$}}& \multirow{2}{*}{\textbf{Avg.}}\\
\cline{3-11}
&&1 &2&3&4&5&6&7&8&9&  \\\hline
Standard Base & ResNet-18 &70.87& 65.71& 61.66& 58.51& 55.49& 52.68& 50.07& 48.08& 46.64 &  56.63 $\textcolor{gray}{(+0.00)}$  \\
CEC~\cite{zhang2021few}&CLIP~\cite{radford2021learning}&77.40&71.94& 67.91&64.69&61.54&58.40&55.46&53.45&52.18& 62.55 $\textcolor{red}{(+5.92)}$ \\
FACT~\cite{zhou2022forward}&CLIP~\cite{radford2021learning}&85.70&80.49&76.11&72.40&68.83&65.55&62.39&60.52&58.66&70.07 $\textcolor{red}{(+13.44)}$ \\
Base-V (Ours)&CLIP~\cite{radford2021learning}& 72.56& 69.63 & 66.38& 62.77 &60.60& 57.33& 54.34& 52.16& 50.49&60.70 $\textcolor{red}{(+4.03)}$ \\
Ours (LRT)&CLIP~\cite{radford2021learning}& \textbf{90.17}& \textbf{85.82} &\textbf{81.70}& \textbf{78.12}& \textbf{75.04}& \textbf{71.71}&\textbf{68.88}& \textbf{66.74}& \textbf{65.34}&\textbf{75.94} $\textcolor{red}{(+19.31)}$\\
\bottomrule
\end{tabular}
}
}
\end{table*}

\begin{figure}[!t]
\begin{center}
\includegraphics[width=1\columnwidth]{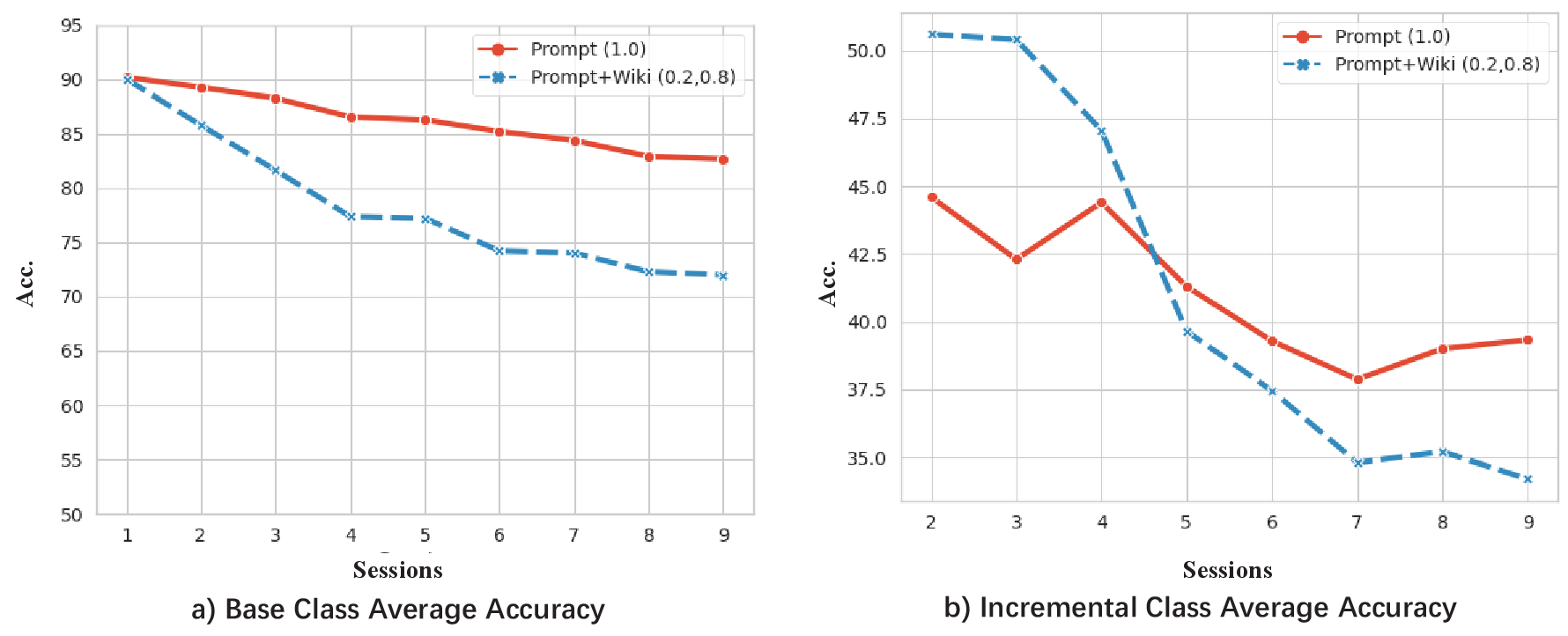}
 \caption{Accuracies of base and the average of our methods and extensions on Wiki data~\cite{bujwid2021large}. Using Wiki data helps the fast understanding of incremental sessions in b), while leading to performance drops on the base sessions in a).
 }\label{fig:wiki}
 \end{center}

\end{figure}

\begin{figure*}[!t]
\begin{center}
\includegraphics[width=0.99\textwidth]{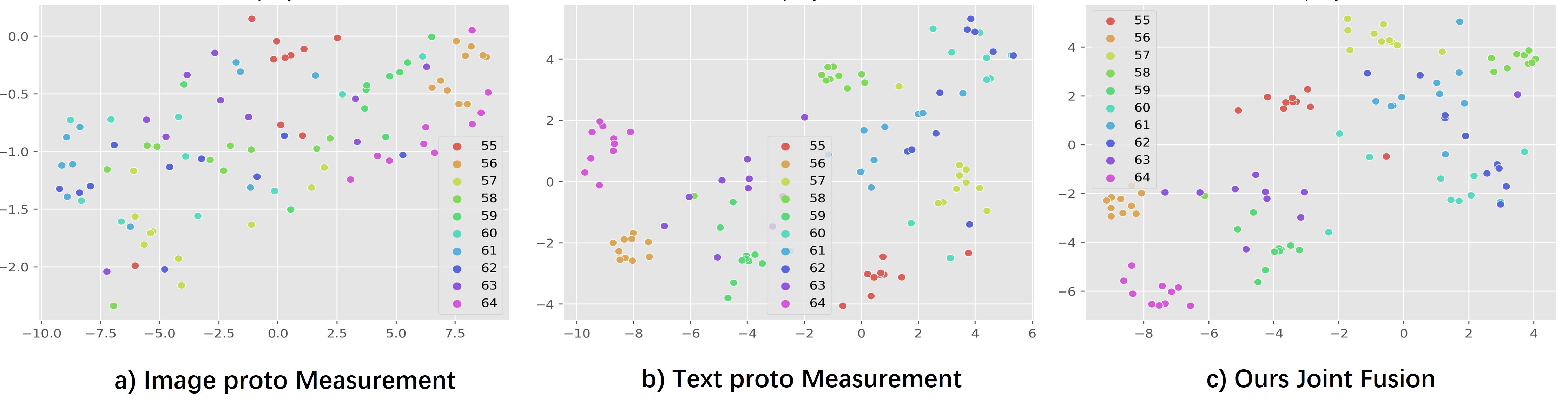}
 \caption{T-SNE visualizations of prediction scores on \textit{mini}ImageNet dataset. a): Measurement only using visual prototypes. b) Measurement only using text prototypes. c) Our proposed joint fusion strategy with text-image measurements. }
 \label{fig:tsne}
 \end{center}
\end{figure*}

\textbf{Understanding Objects from Wiki.} One ideal scenario for understanding novel concepts is learning from descriptions from sufficient web data, which contains rich descriptions like "\textit{The northern cardinal has a distinctive crest on the head...}". To achieve this, we collect the first 5 sentences from the Wikipedia articles corresponding to ImageNet categories provided by~\cite{bujwid2021large}. With this prior knowledge, we fuse the wiki data and our prompt with a ratio of $(8:2)$ with other settings identical.~\figref{fig:wiki} exhibits the results using wiki data (blue dotted line) and our final model (red line). It can be found that the wiki data do provide rich knowledge for incremental classes (\eg, over $8\%$ in session 2) with few learnable samples as in~\figref{fig:wiki} b). However, as prompt learning takes a less important place during this learning, the base classes are less discriminative with the incremental sessions, which leads to clear drops during the base sessions in~\figref{fig:wiki} a). We would like to leave this extension in our future work by incorporating advanced prompt fusion strategies when more text data are available.

\subsection{Discussions}

\textbf{How Does LRT Help FSCIL?} As aforementioned, the major challenge in few-shot class-incremental learning is to alleviate the forgetting of base classes while recognizing novel incremental classes. Some recent research~\cite{peng2022few} indicates that maintaining the base session performance but inferior incremental session performance would also lead to higher results, which are caused by the imbalanced number of classes in base and incremental sessions. Here we conduct detailed comparisons with two state-of-the-art methods~\ie, CEC~\cite{zhang2021few} and FACT~\cite{zhou2022forward} in~\tabref{tab:har}. The \textbf{Base Acc.} denotes the averaged base class accuracy after the final incremental session, and similarly, we define the \textbf{Inc. Acc.} for all incremental classes. We then calculate the harmonic average accuracy (\textbf{Har. Acc.}) of incremental and base accuracy. Our proposed method achieves over $22\%$ improvements in the incremental accuracy on \textit{mini}ImageNet dataset. Moreover, increasing the length of text prompts (from \textbf{2PT} to \textbf{4PT}) leads to a slight performance drop ($1.6\%$)  on base classes, while boosting the incremental accuracy for over $3.9\%$.

Besides the comparison with recent methods, the other natural concern is: \textit{why our proposed LRT improves the representation of incremental knowledge?}
Keeping this in mind, we conduct ablations to verify how the knowledge transfer improves the learning of incremental sessions, as in~\tabref{tab:trans-ablation}. The results indicate that our proposed imagined image-text contrastive loss greatly improves the alignment of text and image domains and thus improves the incremental accuracy,~\ie, from $10.15\%$ to $24.55\%$. With our proposed graph module $\mc{M}_{graph}$ in the last line, the incremental accuracy can be improved to $39.32\%$, which results in a performance margin compared to the prevailing works.

\begin{table}[!t]
\centering{
\caption{ Long-term incremental learning (21 sessions) on  on \textit{mini}ImageNet for 5-way 5-shot classification.}
\label{table:long}
\setlength{\tabcolsep}{0.8mm}
\renewcommand{\arraystretch}{1.4}
\footnotesize
\resizebox{\linewidth}{!}{
\begin{tabular}{l|c|c|c|c|c}
\toprule
Methods &   Acc $(\mc{D}^1)$&  Acc $(\mc{D}^{11})$&  Acc $(\mc{D}^{21})$ & Avg. Acc& $\Delta_{avg}$\\
\midrule
CEC-CLIP~\cite{zhang2021few}&77.40&61.53&52.18&62.40&$\textcolor{gray}{(+0.00)}$\\
FACT-CLIP~\cite{zhou2022forward}&85.70&68.83&58.66&69.94&$\textcolor{red}{(+7.54)}$\\
Ours (LRT)&90.13&73.41&63.64&74.64&$\textcolor{red}{(+12.24)}$\\
\bottomrule
\end{tabular}
}}
\end{table}

\textbf{Do the Performance Improvements Mainly Benefited from the CLIP Pretraining?}
There is no doubt that our proposed model benefits from the language-vision pretraining~\cite{radford2021learning}, which could lead to a performance boost even with the naive training scheme. We thus conduct detailed comparisons with the standard baseline and our base models (\textbf{Base-V}). The standard training baseline is modified from the decoupled prototype learning in~\cite{zhang2021few}. The major difference between these two training procedures is the multi-modality pretraining by CLIP~\cite{radford2021learning}. As in~\tabref{table:base}, the pretrained CLIP model provides a performance gain in both base session and incremental session with an average of $4\%$ on the \textit{mini}ImageNet dataset. The CEC models surpass the baseline visual tuning models with an average of over $1.85\%$ and FACT achieves an average accuracy of $70.97\%$. Under different circumstances, our proposed LRT is able to achieve steady improvements on both base and incremental sessions by a clear margin, which indicates our performance improvements do not mainly come from the strong pretraining models but develop the potential of multimodal knowledge transfer.

\textbf{Can Models Learn from Long-term Incremental Sessions?}
We conduct detailed experimental comparisons with CEC~\cite{zhang2021few} and FACT~\cite{zhou2022forward} with the replaced CLIP backbones. We split the \textit{mini}ImageNet dataset into two parts (base session for 60 classes and incremental session for the rest 40 classes). The incremental stage consists of 20 sessions, where each session has 2 classes $\times$ 5 samples.  The parameters of the network are fixed in CEC and FACT during the incremental session, which makes these methods show less forgetting in the long-term incremental learning but limits their crucial ability to learn novel concepts. Even under this setting, our proposed LRT still shows preferable performance improvements,~\ie, $12.54\%$ compared to CEC~\cite{zhang2021few}. This is mainly because the incremental relationship is pre-learned in the textual encoders of CLIP models, and during the incremental sessions, its relationship is basically stable and does not suffer from catastrophic forgetting.

\textbf{How Does Text Knowledge Help the Joint Embedding?} The crucial idea of our multi-modal learning paradigm is to transfer the text domain knowledge to the image domains. Here we visualize the t-SNE results of the final prediction scores on \textit{mini}ImageNet dataset. To present clear results, we only select the last 5 base classes (54$\sim$59) and the first 5 incremental classes (60$\sim$64) in the same space. We select the first 10 images of each class in~\figref{fig:tsne}. With the only learnable visual prototypes in~\figref{fig:tsne} a), although the base classes can be distinguished before the incremental stages after continuous learning the incremental classes are entangled with the base classes in the feature embedding. While in~\figref{fig:tsne} b), measurements using the text features show a clear distribution with little confusion. By using the joint measurement of text and image prototypes, final results in~\figref{fig:tsne} c) show clear decision boundaries of each class clusters.

\subsection{Limitations and Future Works} 
As the text domain also retains rich knowledge for understanding the object, simply using text prompts with few-shot visual samples still leads to insufficient representations,~\ie, about 40\% accuracy with 5-shot learning on \textit{mini}ImageNet. It is caused by that only the class token $[\bs{CLS}]$ is used for multi-modal alignment. Other methods~\cite{radford2021learning} indicate that using rich hand-crafted prompts may lead to higher performances, including ``a good photo of $[\bs{CLS}]$". Besides, compared to the visual representations with sufficient training samples, using text embedding as prototypes also lead to performance bottleneck, which might be caused by the insufficient description of local visual patterns. 

One possible solution to solve this limitation is to design a dynamic bi-directional learning strategy for visual and text representations. When sufficient training samples are available (\eg, ImageNet), there should be also a re-adjustment of text embedding. In other words, we have only explored the data flow from $\bs{T}\rightarrow \bs{I}$ in this work, while the $\bs{I}\rightarrow \bs{T}$ relations are not fully discovered, which is also a promising direction for many downstream tasks.

\section{Conclusions}\label{sec:conclusion}
In this paper, we make attempts to explore the few-shot class-incremental learning problem from a novel perspective by introducing generalized pertaining language knowledge as learning guidance. To achieve this, our approach proposes a new language-guided relation transfer module and a text-vision prototypical fusion module for joint text-vision representations. Beyond that, to align text with image data in FSCIL, we introduce context prompt learning for fast adaptation during training and an imagined contrastive loss to alleviate the data insufficiency during multi-modal alignment. Experimental results demonstrate that our proposed method surpasses the conventional single-modal methods by a large margin on benchmark datasets.

\ifCLASSOPTIONcompsoc

  \section*{Acknowledgments}
\else

  \section*{Acknowledgment}
\fi

This work is partially supported by grants from the National Natural Science Foundation of China under contracts No. 62132002, No. 62425101, No. 62088102, and No. 62202010.  This work is also supported by the Fundamental Research Funds for the Central Universities.

\ifCLASSOPTIONcaptionsoff
  \newpage
\fi

\balance
\bibliographystyle{IEEEtran}
\bibliography{incfewshot}

\end{document}